\documentclass[letterpaper]{article} 
\usepackage{aaai24}  
\usepackage{times}  
\usepackage{helvet}  
\usepackage{courier}  
\usepackage[hyphens]{url}  
\usepackage{graphicx} 
\usepackage{natbib}  
\usepackage{caption} 
\usepackage{newfloat}
\usepackage{listings}
\DeclareCaptionStyle{ruled}{labelfont=normalfont,labelsep=colon,strut=off} 
\title{LogoStyleFool: Vitiating Video Recognition Systems via Logo Style Transfer}
\author{
    Yuxin Cao\textsuperscript{\rm 1}\equalcontrib, Ziyu Zhao\textsuperscript{\rm 2}\equalcontrib, Xi Xiao\textsuperscript{\rm 1}\thanks{Corresponding author.}, Derui Wang\textsuperscript{\rm 3}, Minhui Xue\textsuperscript{\rm 3}, Jin Lu\textsuperscript{\rm 4}
}
\affiliations{
    \textsuperscript{\rm 1} Shenzhen International Graduate School, Tsinghua University, China \\
    \textsuperscript{\rm 2} Fan Gongxiu Honors College, Beijing University of Technology, China \\
    \textsuperscript{\rm 3} CSIRO's Data61, Australia \\
    \textsuperscript{\rm 4} Ping An Technology (Shenzhen) Co., Ltd., China \\
}

\usepackage{bibentry}

\usepackage{tikz}
\usepackage{amsmath}
\usepackage{filecontents}
\usepackage{threeparttable}
\usepackage[utf8]{inputenc} 
\usepackage[T1]{fontenc}    
\usepackage{url}            
\usepackage{booktabs}       
\usepackage{amsfonts}       
\usepackage{nicefrac}       
\usepackage{microtype}      
\usepackage{xcolor}         
\usepackage{graphicx}
\usepackage{amssymb}
\usepackage{times}
\usepackage{epsfig}
\usepackage{enumitem}
\usepackage{multirow}
\usepackage{makecell}
\usepackage{dsfont}

\DeclareMathOperator{\findstyle}{\texttt{find\_style}}
\DeclareMathOperator{\logotransfer}{\texttt{logo\_transfer}}
\DeclareMathOperator{\calmask}{\texttt{cal\_mask}}
\DeclareMathOperator{\pad}{\texttt{pad}}
\DeclareMathOperator{\logosdct}{\texttt{LogoS\_DCT}}
\DeclareMathOperator{\update}{\texttt{update}}

\newtheorem{theorem}{Theorem}
\newtheorem{definition}{Definition}
\newtheorem{proposition}{Proposition}
\newtheorem{lemma}{Lemma}

\allowdisplaybreaks[3]

\usepackage[ruled,vlined,linesnumbered]{algorithm2e}
\usepackage{algorithm2e}
\usepackage{algorithmicx}  
\usepackage{algpseudocode}



\usepackage{siunitx}
\def\eg{\emph{e.g.,}\xspace}
\def\etc{\emph{etc}\xspace}
\def\ie{\emph{i.e.,}\xspace}

\begin{document}

\maketitle

\begin{abstract}
Video recognition systems are vulnerable to adversarial examples. Recent studies show that style transfer-based and patch-based unrestricted perturbations can effectively improve attack efficiency. These attacks, however, face two main challenges: 1) Adding large stylized perturbations to all pixels reduces the naturalness of the video and such perturbations can be easily detected. 2) Patch-based video attacks are not extensible to targeted attacks due to the limited search space of reinforcement learning that has been widely used in video attacks recently. In this paper, we focus on the video black-box setting and propose a novel attack framework named \emph{LogoStyleFool} by adding a stylized logo to the clean video. We separate the attack into three stages: style reference selection, reinforcement-learning-based logo style transfer, and perturbation optimization. We solve the first challenge by scaling down the perturbation range to a regional logo, while the second challenge is addressed by complementing an optimization stage after reinforcement learning. Experimental results substantiate the overall superiority of LogoStyleFool over three state-of-the-art patch-based attacks in terms of attack performance and semantic preservation. Meanwhile, LogoStyleFool still maintains its performance against two existing patch-based defense methods. We believe that our research is beneficial in increasing the attention of the security community to such subregional style transfer attacks.
\end{abstract}

\section{Introduction}
Short videos have become omnipresent in the current era. With the tentacles of Deep Neural Networks (DNNs) extending from images to videos, the quality in services such as video recognition~\cite{ji20123d,carreira2017i3d,hu2023overview}, video segmentation~\cite{zhou2022survey,gao2023deep} and video compression~\cite{chen2017deepcoder,ma2019image} has been greatly improved. However, they are also encountered with severe security threats -- DNNs are vulnerable to adversarial examples which are generated by surreptitiously introducing minuscule perturbations fooling the model~\cite{szegedy2014intriguing,goodfellow2015explaining}. For example, attackers can perturb malicious / toxic videos to evade DNN-based detectors, which may result in severe social-economic consequences. Therefore, the outpouring of adversarial videos has raised security-critical concerns for machine-learning-assisted systems such as video verification systems~\cite{cao2023stylefool}. To date, most attacks against videos focus on crafting imperceptible perturbations restricted by $\ell_p$ norms. This branch of attacks considers restricted perturbations and consumes a large number of queries under the black-box setting. A recent work utilizes style transfer to introduce pixel-wise unrestricted perturbations that do not affect the semantic information of the video~\cite{cao2023stylefool}. However, seldom work utilizes sub-region perturbations, such as adversarial patches~\cite{yang2020patchattack,chen2022attacking}, in the attack. Nevertheless, the sub-region perturbations are considered as more generalizable and practical in the physical world.

In this paper, we investigate the risk brought by unrestricted sub-region perturbations towards video recognition systems. Such attacks face two main challenges. 1) Style transfer-based attacks add perturbations to all pixels~\cite{cao2023stylefool}, which may cause artifacts as a result of naturalness reduction. Such perturbations can also be detected with ease~\cite{xiao2019advit,jia2019comdefend}. 2) Due to the limited search space of reinforcement learning (RL), patch-based video attacks~\cite{chen2022attacking} only support untargeted attacks (the performance of the targeted version plummets), while targeted attacks are in higher demand and of greater harm. From another perspective, despite many existing patch-based attacks in the image domain~\cite{brown2017adversarial,karmon2018lavan,chindaudom2020adversarialqr,camera_li2019adversarial,yang2020patchattack,gong2023stealthy}, directly extending patch-based attacks from images to videos is hard due to the increase in data dimensionality and the lack of temporal consistency.

To address the above challenges, we follow the paths of both style transfer-based attacks and patch-based attacks in the video domain and propose \textbf{LogoStyleFool}, a black-box adversarial attack against video recognition systems via logo style transfer. Dissimilar to the existing style transfer-based video attack which imposes perturbations on all pixels, we first plumb the possibility of vitiating video recognition systems by transferring the style of the logo and forming local perturbations to superimpose on the video. Our attack first finds style images with random color initialization and builds a style set based on the intuition that the style image, which can be classified as the target class, carries more information about the target class, thus facilitating the attack efficiency. Then the best logo, style image and position parameters are searched by RL, and the adversarial video is initiated by the original video regionally superimposed with the stylized logo. In contrast to irregular patches, the stylized logos that we integrate into the content encompass valuable semantic information, thereby preserving the inherent naturalness of the video. Additionally, our approach facilitates a meticulous logo placement, striving to position the logo in proximity to the video's corners while suppressing its size. This optimization is achieved through RL, aiming at maximizing the overall visual naturalness. Finally, the adversarial video is updated through iterative optimizations, which alleviates the limited search space of RL in existing methods. We prove the upper bound of both $\ell_\infty$ and $\ell_2$ partial perturbations for videos in the perturbation optimization process. Through experiments compared with three existing patch-based attacks, LogoStyleFool can launch both targeted and untargeted attacks and achieve better attack performance and semantic preservation. Moreover, we demonstrate the performance of LogoStyleFool against two existing patch-based defense methods. In summary, our contributions are listed as follows.

\begin{itemize}
    \item We propose a brand new attack framework, termed LogoStyleFool, which superimposes a stylized logo on the input video, against video recognition systems. LogoStyleFool sets up a holistic approach to patch-based attacks.
    \item We provide a better action space in style reference selection and initialize the video by RL-based logo style transfer, which can move the video with a stylized logo close to the decision boundary and improve the attack efficiency. We also design a novel reward function that considers the distance between logos and the corners of the video to ensure their naturalness.
    \item We also complement a perturbation optimization stage after RL to solve the problem of limited search space widely present in the existing {patch/}RL-based attacks, making {patch/}RL-based attacks extensible to targeted attacks. The upper bounds of both the $\ell_\infty$ and $\ell_2$ partial perturbations assure the video's naturalness and temporal consistency.
    \item We show that LogoStyleFool can achieve superior attack performance and preserve semantic information in both targeted and untargeted attacks while maintaining the performance against patch-based defense methods.
\end{itemize}

\section{Related Work}
\noindent\textbf{Deep-Learning-Based Video Adversarial Attacks.}
In the early stage, attacks are launched under the white-box setting, where the attacker can access the model architecture and parameters~\cite{wei2019sparse,li2019stealthy,inkawhich2018adversarial,pony2021flickering,chen2021appending}. For commercial systems, it is hard to secure the inner information of the model, thus, black-box attacks are more practical. Black-box video adversarial attacks assume that only the top-1 score and its label are available to attackers. Jiang et al.~\cite{jiang2019black} first proposed V-BAD to dupe video recognition systems by introducing tentative perturbations. Wei et al.~\cite{wei2020heuristic} proposed a heuristic attack to add sparse perturbations to the input sample both temporally and spatially. Cao et al.~\cite{cao2023stylefool} first used style transfer to add unrestricted perturbations to video samples and reduce queries by a large margin. However, perturbations in all pixels may bring about local artifacts (\eg green skin, blue leaves) that affect the naturalness.

\noindent\textbf{Reinforcement-Learning-Based Adversarial Attacks.}
Deep RL was originally designed to learn and simulate human decision-making processes. RL has received a lot of attention, including those engaged in the issue of adversarial attacks. Yang et al.~\cite{yang2020patchattack} proposed a patch-based black-box image attack method, PatchAttack, which utilizes RL to optimize the location and texture parameters of each patch to generate adversarial samples. Wei et al.~\cite{wei2022sparse} first applied RL to black-box video attacks by designing an agent based on attack interaction and intrinsic attributes of the video to select the keyframes of the video. Similarly, RL has also been used in key frame/region selection~\cite{wang2021reinforcement,wei2023efficient} and perturbation optimization~\cite{yan2021efficient}. Reinforcement-learning-based attacks are encountered with the problem of limited search space, making most of them unadaptable to targeted attacks.

\noindent\textbf{Patch-Based Attacks.}
In the image domain, Jia et al.~\cite{jia2020adv} proposed Adv-watermark, which fooled classifiers by adding meaningful watermarks such as school badges and trademarks. However, the transparency of watermarks weakens the attack capability when the perturbed area is limited and is not effective when attacking higher dimensional data such as videos. Croce et al.~\cite{croce2022sparse} proposed a versatile framework, Sparse-RS. It incorporates a random color patch method, but tends to yield results that exhibit less naturalness. To the best of our knowledge, BSC~\cite{chen2022attacking} is the first patch-based attack in the video domain by adding bullet-screen comments. Although BSC maintains the naturalness of the video to some extent and slightly reduces queries, it can only launch untargeted attacks due to the problem of limited search space. A concurrent work~\cite{jiang2023efficient} proposes an efficient decision-based patch attack for videos using a spatial-temporal differential evolution framework (STDE). However, patches in targeted attacks are large enough (as discussed later) to affect the semantic information of clean videos. To conclude, there is a trade-off between attack efficiency and perturbation stealthiness.

\begin{figure*}[t]
\begin{center}
  \includegraphics[width=0.7\linewidth]{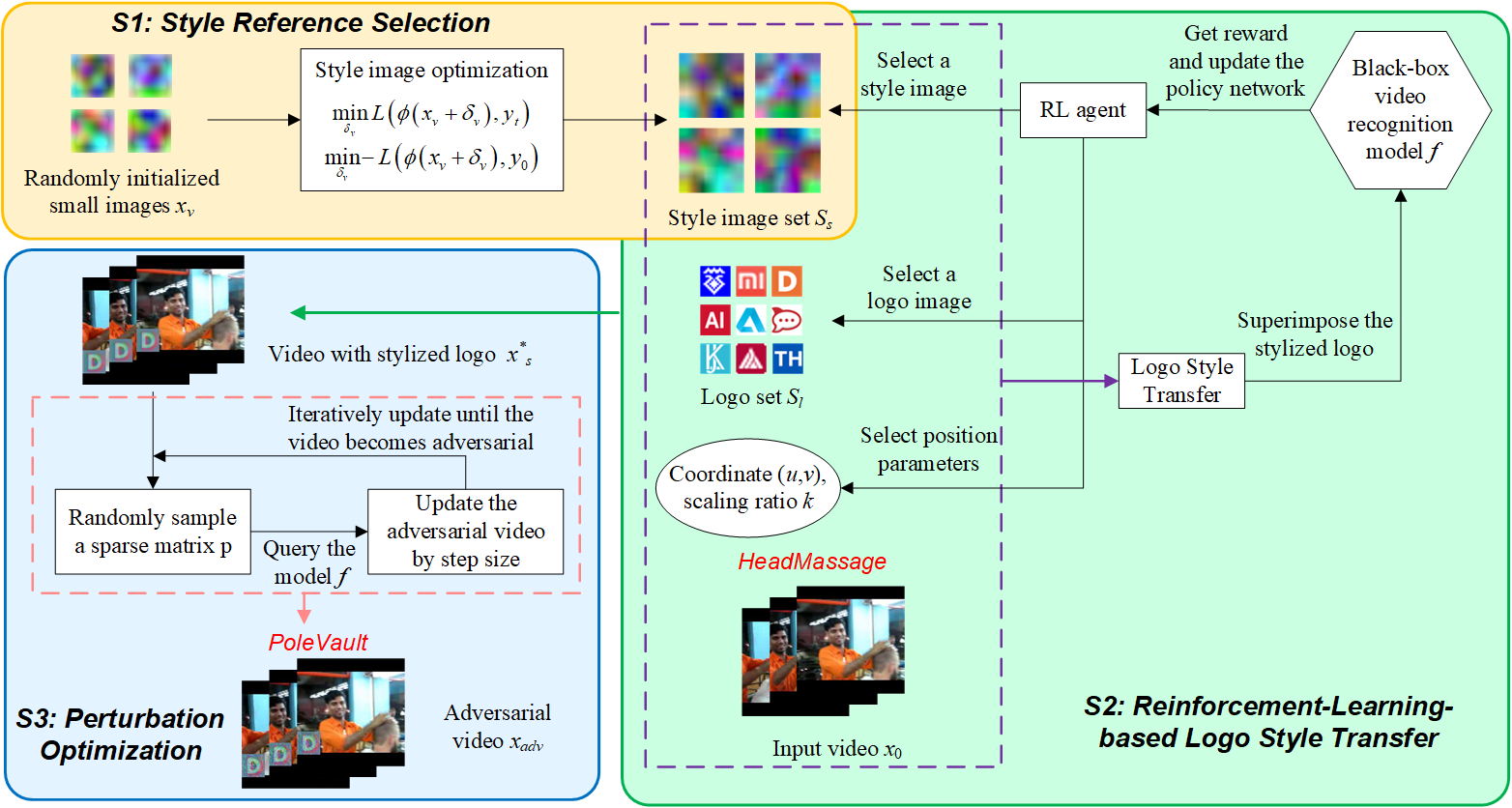}
\end{center}
  \caption{Overview of our proposed LogoStyleFool.}
\label{fig:framework}
\end{figure*}

\section{Methodology}
For the target model $f$, it takes a video $x\in {\mathbb{R}^{T \times H \times W \times C}}$ as input, and outputs its predicted label $y$ and its score $p \left(y|x \right)$, where $T$, $H$, $W$ and $C$ respectively represent the frame number, height, width and channel number of the video. The attacking goal is to find a meticulously fabricated logo in a certain style and superimpose it on a certain position of the video, so as to fool the target model, \ie $f\left( x_{adv} \right) \ne y_0$ for untargeted attacks and $f\left( x_{adv} \right) = y_t$ for targeted attacks, where $x_{adv}$ presents the adversarial video, $y_0$ and $y_t$ denote the original label and the target label respectively. We consider the black-box setting where the attacker can only access the top-1 score and its label. We also assume that the target model has a query limit. The framework of LogoStyleFool is depicted in Figure~\ref{fig:framework}. We separate LogoStyleFool into three stages: style reference selection, reinforcement-learning-based logo style transfer, and perturbation optimization.

\subsection{Style Reference Selection}
In order to make the stylized logo carry more information about the target label for targeted attacks (the other labels for untargeted attacks), and to reduce computational complexity, we use block perturbations to search for style images. Concretely, for a small image $x_v \in {\mathbb{R}^{{H_b} \times {W_b} \times C}}$, where $H_b$ and $W_b$ represent the block height and width respectively, the goal is to find the optimal block perturbations ${{\delta _v}}$ that satisfy
\begin{small}
\begin{equation}
\left\{ \begin{array}{l}
\mathop {\min }\limits_{{\delta _v}} L_{CE}\left( {\phi \left( {{x_v} + {\delta _v}} \right),{y_t}} \right), ~~ \rm{targeted},\\
\mathop {\min }\limits_{{\delta _v}}  - L_{CE}\left( {\phi \left( {{x_v} + {\delta _v}} \right),{y_0}} \right), ~~ \rm{untargeted},
\end{array} \right.
\end{equation}
\end{small}%
where $L_{CE}$ represents the cross-entropy loss, $\phi$ represents the resize operation that expands the small perturbed image to the same dimension as the input video $x$. We use a simple black-box attack, SimBA~\cite{guo2019simple}, to optimize the perturbations and do not restrict the $\ell_p$ norm of perturbations ${{\delta _v}}$ here since the naturalness of the style image is not of significance. We denote the style image as ${x_v^*} = {x_v + \delta _v^*}$ after obtaining the optimal block perturbations $\delta _v^*$. 

As the initial values of $x_v$ affect the style image, we randomly initialize $x_v$ multiple times to obtain $N_s$ style images, which also provides a larger search space for subsequent reinforcement learning. The set consisting of different style images is called the style set $S_s$ for an input video.

\subsection{Reinforcement-Learning-Based Logo Style Transfer}
\textbf{Logo Style Transfer.}
We perform style transfer for the logo region to generate more natural but unseen logos. Such a process can help conceal adversarial perturbations and meanwhile increase the difficulty for adversarial detection, since the adversarial samples are diversified when using different style images.
With the aid of traditional image style transfer method~\cite{gatys2015a,justin2016perceptual,li2018learning}, we consider content loss $L_{cont}$, style loss $L_{sty}$, and total variance loss $L_{tv}$ for logo image $l$ and style image $x_v^*$. The stylized logo $l_s^*$ can be expressed as
\begin{small}
\begin{equation}
\label{equation:total_loss}
l_s^* = \mathop {\arg \min }\limits_{{l_s}} {\lambda _c}{L_{cont}}\left( {l,{l_s}} \right) + {\lambda _s}{L_{sty}}\left( {{l_s},{x_v^*}} \right) + {\lambda _{tv}}{L_{tv}}\left( {{l_s}} \right),
\end{equation}
\end{small}%
where $\lambda _c$, $\lambda _s$ and $\lambda _{tv}$ represent weight coefficients. For more details about style transfer losses, please refer to the work done by Justin et al.~\cite{justin2016perceptual}.

\noindent\textbf{Logo Set.}
LLD~\cite{sage2018logo} is a publicly available large logo dataset that includes over 600,000 logos worldwide. Some logos have a large number of transparent pixels, which may, after style transfer, cause abnormal perturbations (irregular colors on the white background) and weaken the concealment of stylized logos. Therefore, these logos are not suitable for style transfer. We automatically filter the logos in the dataset, remove logos with transparent pixels or a large number of white pixels, and then randomly select $N_l$ logos to construct the desired logo set $S_l$.

\noindent\textbf{Reinforcement Learning.}
Choosing the appropriate location, logo, and style for the video is a key step for attacks. Specifically, define the search space $S$ as a set of $ \left( u, v, k, l_{ind}, s_{ind} \right)$ with dimension $D_{S}$, where $\left( u, v \right)$ is the pixel coordinate of the upper left corner of the logo region, $k$ is the scaling ratio of the logo, $l_{ind}$ is the index of the logo selected from the logo set and $s_{ind}$ is the index of the style selected from the style set. Given the height $h$ and width $w$ of the logo, then $u \in \left[ 0, H - kh \right] \cap \mathbb{Z}$, $v \in \left[ 0, W - kw \right] \cap \mathbb{Z}$. The training agent takes 5 actions in sequence to generate an action sequence $a \in S$, \ie selecting the logo position, scaling size, logo and style. Then the stylized logo is resized and superimposed on the video position.

We raise three requirements for stylized logos. Firstly, videos with stylized logos can be moved as close as possible to the decision boundary of the target label for targeted attacks (the original label for untargeted attacks). Secondly, the logo area should be small, making it less prominent in the video. Thirdly, the logo should be located in the corners of the video, which is not enough to affect the core semantic part of the video. Thus, the reward function is defined as
\begin{small}
\begin{equation}
R = \left\{ \begin{array}{l}
\log p\left( {{y_t}|{x_s}} \right) - {\mu _a}{k^2}hw - {\mu _d}d_m, ~~ \rm{targeted},\\
\log \left( {1 - p\left( {{y_0}|x_s} \right)} \right) - {\mu _a}{k^2}hw - {\mu _d}d_m, ~~ \rm{untargeted},
\end{array} \right.
\end{equation}
\end{small}%
where $x_s$ denotes the video superimposed with a stylized logo, $d_m = \min \left\{ {{d_1},{d_2},{d_3},{d_4}} \right\}$ denotes the shortest distance from the four corners of the logo to the nearest corner in the video, ${d_1} = {\left\| {\left( {u,v} \right) - \left( {0,0} \right)} \right\|_2}$, ${d_2} = {\left\| {\left( {u,v + kw} \right) - \left( {0,W} \right)} \right\|_2}$, ${d_3} = {\left\| {\left( {u + kh,v} \right) - \left( {H,0} \right)} \right\|_2}$, ${d_4} = {\left\| {\left( {u + kh,v + kw} \right) - \left( {H,W} \right)} \right\|_2}$, $\mu_a$ and $\mu_d$ are the area and distance coefficients balancing the penalties on the logo size and the distance from the logo to the corner.

Following prior work~\cite{yang2020patchattack,chen2022attacking}, the policy network, which is composed of an LSTM and a fully-connected layer, outputs the probability distribution of actions $p\left( {{a_t}|\left( {{a_1},\; \cdots ,{a_{t - 1}}} \right)} \right)$ in step $t$, the action of which $a_t$ is later sampled from the Categorical function. After traversing all actions, the agent will generate a sampling probability ${\pi _{{\theta _p}}}$ and a reward function $R$ for an action sequence. Since the purpose of the policy network is to minimize the distance between the sampling distribution and the reward, the loss function is defined as $L\left( {{\theta _p}} \right) =  - {\mathbb{E}_{\tau  \sim {\pi _{{\theta _p}}}}}\left[ {R\left( \tau  \right)} \right]$. The policy network parameter ${\theta _p}$ can be updated by calculating the approximated gradient after multiple trajectory sampling~\cite{williams1992simple}.
\begin{small}
\begin{equation}
{\nabla _{{\theta _p}}}L\left( {{\theta _p}} \right) \approx - \frac{1}{\Omega }\sum\limits_{\tau  = 1}^\Omega  {\sum\limits_{t = 1}^{D_{S}} {{\nabla _{{\theta _p}}}\log {\pi _{{\theta _p}}}\left( {{a_t}|{h_t}} \right)} } R\left( \tau  \right),
\end{equation}
\end{small}%
where $\Omega$ denotes the number of sampling trajectories, $h_t$ denotes the hidden state of LSTM. By continuously optimizing action sequences through agents, a set of optimal action sequences can be obtained, which can move videos added with a stylized logo towards the decision boundary.

\subsection{Perturbation Optimization}
SimBA-DCT~\cite{guo2019simple}, the DCT (Discrete Cosine Transform) version of SimBA, is a query-efficient attack in the image domain. However, squarely extending SimBA-DCT to videos faces challenges. The video dimension is much larger than that of images so the attack cannot succeed if we just change every channel by step size only once. On the other hand, the attack difficulty surges if we perturb only a small region in the video. To solve these obstacles, we improve SimBA-DCT to optimize the perturbations for the logo region after reinforcement learning and name it LogoS-DCT. Optimization becomes the following.
\begin{small}
\begin{equation}
\mathop {\min }\limits_\delta  L_{CE}\left( {x_s^* + M \odot \delta ,{y_t}} \right),{\rm{s}}{\rm{.t}}{\rm{.}}{\left\| {M \odot \delta } \right\|_p} \le \varepsilon ,
\end{equation}
\end{small}%
where $x_s^*$ represents the video superimposed with the optimized logo after reinforcement learning, $M\in {\mathbb{R}^{T \times H \times W \times C}}$ denotes the logo mask where the value in the logo region is 1, otherwise 0, $\delta$ represents the perturbation, $\varepsilon$ stands for the perturbation threshold.

Following SimBA-DCT~\cite{guo2019simple}, we iteratively optimize the perturbation by randomly sampling from the orthogonal frequency set $Q_{DCT}$ extracted by DCT. According to previous experimental attempts, most videos cannot be attacked in one round for targeted attacks, which is due to the fact that the dimension of videos is much larger than images. To address this challenge, we modify the perturbation update strategy. Attackers can perform multiple rounds of modification on all pixels in the input sample until the attack succeeds, but each pixel in each round can only be changed by one of $\left\{ - \eta, 0, \eta \right\}$ compared to the initial pixel. $\eta$ denotes the step size.

\begin{proposition}\label{proposition1}
The perturbation after $K$ steps can be expressed as 
\begin{small}
\begin{equation}
M \odot {\delta _K} = M \odot \sum\limits_{m = 1}^{\min \left\{ {K,d} \right\}} {\Psi \left( {{A^T}\left( {{\gamma _m}\eta {{\rm{p}}_m}} \right)A} \right)} ,
\end{equation}
\end{small}%
where $\Psi \left( x \right) = {\rm{clip}}_{ - \varepsilon }^{ + \varepsilon }\left( x \right)$ for $\ell_\infty$ restriction, $x$ for $\ell_2$ restriction, $\rm{clip}$ stands for the clip operation to restrict the perturbation within the $\ell_\infty$ ball. $A$ represents the DCT transformation matrix, which satisfies ${A_{ij}} = c\left( i \right)\cos \left[ {\frac{{\left( {j + 0.5} \right)\pi }}{d}i} \right]$, where $c\left( i \right)$ equals to $\sqrt {\frac{1}{d}}$ if $i=0$, $\sqrt {\frac{2}{d}}$ otherwise. $d$ denotes the dimension after flattening the video. ${{\gamma _m}} \in \left\{ { -1,0,1 } \right\}$ denotes the product of signs in each round in the $m$-th pixel. $\rm{p}_m$ is a sparse matrix where only one element is 1, which indicates that only one channel in a certain pixel is considered at a time to optimize the direction. 
\end{proposition}

Given Proposition~\ref{proposition1}, we can derive the upper bound of perturbation for videos as follows, which means that the perturbation optimized by LogoS-DCT will not significantly break the naturalness of the video with a stylized logo.

\begin{theorem}\label{theorem1}
The $\ell_\infty$ norm of the video adversarial perturbations in the logo area is upper bounded by $\varepsilon \rho \sqrt {\min \left\{ {K,d} \right\}} $, where $\rho  = k\sqrt {\frac{{hw}}{{HW}}} $ signifies the square root of the ratio of the logo area to a frame image.
\end{theorem}

The upper bound of the perturbation can be extended to $\ell_2$ restrictions due to the peculiarity of DCT transformation matrix $A$ and the orthogonality of sparse matrix $\rm{p}_m$.
\begin{lemma}\label{lemma1}
The $\ell_2$ norm of any row in the DCT transformation matrix $A$ is 1.
\end{lemma}

\begin{theorem}\label{theorem2}
The $\ell_2$ norm of the video adversarial perturbations in the logo area is upper bounded by $\eta \rho \sqrt {\min \left\{ {K,d} \right\}} $.
\end{theorem}

Therefore, the upper bound of the perturbation is guaranteed. Proofs are given in the supplement. It is obvious that the increase in either the step size (for $\ell_2$ norm), the logo size or the step number will result in a decrease in the other two. Consequently, it is necessary to find a dynamic balance among the three parameters mentioned above, \ie reasonably controlling the step size and logo size, to reduce the query number (reflected by step number) without affecting visual perception. We provide results for attacks restricted by $\ell_2$ and $\ell_{\infty}$ norms, respectively, in the experimental section.

\subsection{LogoStyleFool Recap}
To sum up, LogoStyleFool can be separated into three stages. Firstly, the style set is built by finding multiple style images that can be misclassified. Then the optimal combination of style index, logo index, position, and size is selected through RL to obtain the video superimposed with the best stylized logo which is close enough to the decision boundary. Finally, the perturbation in the logo area is ulteriorly optimized through LogoS-DCT to obtain the adversarial video. The overall process of LogoStyleFool is shown in Algorithm~\ref{alg:logostylefool}. $\findstyle$ outputs the style set where the images are adversarial and obtained through random initialization and unrestricted SimBA optimization. $\logotransfer$ denotes the style transfer for the logo image. $\calmask$ outputs a mask matrix according to the logo position and size. $\pad$ represents scaling the logo image to a specific position in a frame, and padding pixels outside the logo with 0. The concrete process of LogoS-DCT is provided in the supplement. The source code is available at \url{https://github.com/ziyuzhao-zzy/LogoStyleFool}.

\noindent
\begin{algorithm}[t]
\caption{LogoStyleFool.}\label{alg:logostylefool}\small
\KwIn{Black-box classifier $f$, input video ${x_0}$, original label $y_0$, target label ${y_t}$, style image number $N_s$, logo set $S_l$, search space $S$, orthogonal frequency set $Q_{DCT}$, step size $\eta$, perturbation threshold ${\varepsilon}$.}
\KwOut{Adversarial video ${x_{adv}}$.}
$S_s \gets \findstyle(f, N_s, y_t )$; // replace $y_t$ with $y_0$ for untargeted \\
\While {not meeting the termination condition}{
    ${a} \leftarrow$ an action sequence $\left( {u,v,k,{l_{ind}},{s_{ind}}} \right)$ sampled from $S$\;
    $x_v^* \leftarrow $ $S_s \left( s_{ind}\right)$\;
    $l_s^* \leftarrow \logotransfer(S_l \left( l_{ind}\right), x_v^*)$\;
    $M \leftarrow \calmask \left(u, v, k, x_0\right)$\;
    $x_s \leftarrow x_0 + M \odot \pad\left( {l_s^*} \right)$\;
    Calculate reward $R$ for targeted/untargeted attacks\;
    Calculate RL loss gradient ${\nabla _{{\theta _p}}}L\left( {{\theta _p}} \right)$\;
    Update policy network and the best video $x_s^*$\;
    }
$x_{adv} \gets \logosdct \left( f, x_s^*, y_t, M, Q_{DCT}, \eta, \varepsilon \right)$. // replace $y_t$ with $y_0$ for untargeted
\end{algorithm}

\section{Experiments}
\subsection{Experimental Setup}\label{subsec:setup}
\noindent\textbf{Datasets and Models.}
We choose UCF-101~\cite{soomro2012ucf101} and HMDB-51~\cite{kuehne2011hmdb}, two datasets that are popularly used in video adversarial attacks, to verify the attack performance. We select two frequently used video recognition models, C3D~\cite{tran2015learning} and I3D~\cite{carreira2017i3d}, as our target models. We beforehand trained the two models on two datasets. The video recognition accuracy for C3D and I3D on UCF-101 is 83.54\% and 61.70\%, while that on HMDB-51 is 66.77\% and 47.92\%. Please refer to the supplement for more introduction.

\noindent\textbf{Benchmarks.}
We choose PatchAttack~\cite{yang2020patchattack}, BSC~\cite{chen2022attacking} and Adv-watermark~\cite{jia2020adv} as benchmarks. We extend PatchAttack to videos and consider the rectangular patch with RGB perturbations for a fair comparison. Since BSC only provides results of untargeted attacks, we slightly modify it to adapt to targeted attacks. Both benchmarks optimize the patch iteratively by RL, and the attack is early stopped once the reward converges. However, it is not guaranteed that the attack has succeeded, especially for targeted attacks. Since we consider the query limit in our attack, we enlarge the batch size and the iteration step for both benchmarks to achieve comparative fairness. Owing to the inherent resemblance between watermarks and logos, we extend the application of Adv-watermark~\cite{jia2020adv} to video attacks and compare it as a benchmark to our method. Following two existing video attacks V-BAD~\cite{jiang2019black} and StyleFool~\cite{cao2023stylefool}, we set the query limit as $3 \times 10^5$. The other parameters of benchmarks are set as their default values. As the positions of the bullet-screen comments in the video vary across frames in the BSC attack, the application of our Stage 3 (\ie perturbation optimization) becomes less feasible. To ensure a fair comparison, we provide outcomes obtained through our method without Stage 3.

\noindent\textbf{Metrics.}
We use the following metrics to evaluate the attack performance. 1) Fooling rate (FR) and first two-stage fooling Rate ($^2$FR). 2) Average query (AQ), first two-stage average query ($^2$AQ), average query in each stage (AQ$_1$, AQ$_2$, and AQ$_3$). 3) Average Occluded Area (AOA). 4) Temporal Inconsistency (TI)~\cite{lei2020blind}. We leave the definition of metrics, parameter settings and germane analyses in the supplement.

\begin{table*}[t]
\centering
\footnotesize
\begin{tabular}{ccrrrrrrrrrrrrrrrr}
\toprule
\multirow{2}{*}[-0.5ex]{Model} & 
\multirow{2}{*}[-0.5ex]{Attack} & \multicolumn{4}{c}{UCF-101-Targeted} &  \multicolumn{4}{c}{UCF-101-Untargeted} \\
\cmidrule(r){3-6}\cmidrule(r){7-10}
& & FR($^2$FR)$\uparrow$ & AQ($^2$AQ)$\downarrow$ & AOA$\downarrow$ & TI$\downarrow$ & FR($^2$FR)$\uparrow$ & AQ($^2$AQ)$\downarrow$ & AOA$\downarrow$ & TI$\downarrow$ \\
\midrule
\multirow{5}{*}{C3D} 
& Adv-watermark & 2\% & 824.3 & \textbf{4.38\%} & 5.26 & 46\% &  182.1 & \textbf{4.36\%} & \textbf{4.26} \\
& PatchAttack & 6\% & 37,562.5& 6.32\% & 65.35 & 71\% & 7,004.7 & 6.81\% & 73.53 \\
& BSC & 16\% & 32,886.3& 6.29\% & 5.07 & 83\% & 4,611.8 & 7.50\% & 4.96 \\
& LogoStyleFool-${\ell_\infty}$ & 49\%(9\%) & 26,382.4(2,710.5)& 5.15\% & 4.24 & 97\%(\textbf{81\%})& \textbf{3,308.9}(996.7) & 6.02\% & 4.32 \\
& LogoStyleFool-${\ell_2}$ & \textbf{58\%}(8\%) & \textbf{26,003.4}(1,893.1) & 5.16\% & \textbf{4.16} & \textbf{98\%}(79\%) & 3,463.0(\textbf{933.9})& 5.85\% & 4.48 \\
\midrule
\multirow{5}{*}{I3D} 
& Adv-watermark & 1\% & 876.0 & 5.02\% & 5.72 & 48\% & 571.8 & \textbf{4.66\%} & 4.05 \\
& PatchAttack & 2\% & 31,805.3 & 6.23\% & 34.67 & 66\% & 2,515.7 & 5.74\% & 25.86\\
& BSC & 14\% & 33,517.2 & 7.01\% & \textbf{3.19} & 82\% & \textbf{2,018.0} & 6.60\% & 4.22\\
& LogoStyleFool-${\ell_\infty}$ & \textbf{42\%}(6\%) & \textbf{22,856.3}(2,378.6)& \textbf{4.89\%} & 3.51 & \textbf{97\%}(\textbf{85\%}) & 2,279.1(\textbf{680.3})& 5.84\% & \textbf{3.60} \\
& LogoStyleFool-${\ell_2}$ & 31\%(5\%) & 33,013.0(1,441.2)& 5.06\% & 3.67 & 92\%(80\%) & 3,742.9(741.4)& 5.88\% & 3.65 \\
\bottomrule
\end{tabular}
\caption{Attack performance comparison on UCF-101. Metric details are provided in the experimental setup.}
\label{tab:attack_performance_ucf101}
\end{table*}

\subsection{Experimental Results}\label{subsec:experimental_results}
\noindent\textbf{Attack Performance.}
We randomly select 100 videos respectively from UCF-101 and HMDB-51 to attack C3D and I3D. These videos are all correctly classified as their ground-truth labels. Table~\ref{tab:attack_performance_ucf101} and Table~\ref{tab:attack_performance_hmdb51} in the supplement report the attack performance among 4 attack frameworks (we provide both $\ell_\infty$ and $\ell_2$ versions for LogoStyleFool). Results show that although LogoStyleFool does not exhibit a dramatic (yet still pretty good) edge over PatchAttack and BSC in terms of AQ for targeted attacks, the FR of LogoStyleFool increases a lot. Due to the dimension gap between images and videos and the different attack capabilities between watermark and patch caused by transparency, Adv-watermark performs the worst, followed by PatchAttack. While the query count in Adv-watermark remains relatively modest, the corresponding success rate is notably low, with the majority of samples converging with a high loss before reaching the upper query limit. As for BSC, we find that merely increasing batch size and iteration step can increase the FR somewhat, but the attack performance is still limited due to limited search space, the problem of which has not been fundamentally resolved. We discover that the score of the target class is very low (usually below the power of 10e-3) when the reward converges. We deduce that this issue is not obvious in untargeted attacks attributed to the lower difficulty of untargeted attacks. The $^2$FR and $^2$AQ of LogoStyleFool also support this conjecture, since LogoStyleFool achieves comparable $^2$FR if the attack only has the first two stages (RL in Stage 2). From another respect, increasing search space may intuitively increase the fooling rate of RL-based methods, but we find that increasing batch size cannot significantly improve FR, but instead increases AQ. This indicates that RL may not find an adversarial example that can be misclassified to a certain target class even if the search space is large enough. We address the above issues by adding perturbation optimization after RL in LogoStyleFool, resulting in better attack performance. Though the LogoS-DCT process is the most query-consuming process in three stages, we reduce queries in the first two stages by initializing better style images and stylized logos. Thus, LogoStyleFool can greatly improve the fooling rate with comparable or even less AQ compared with PatchAttack and BSC in targeted attacks. Of course, there are some instances where, despite the stylized logo carrying the target class's information, the original video is too far from the target class's decision boundary, leading to a failed attack due to the query limit. We will try to solve this problem in the future. The difference between LogoStyleFool-${\ell_\infty}$ and LogoStyleFool-${\ell_2}$ is reflected in Stage 3. The average $\ell_2$ distance before and after Stage 3 is only 10.89 (1.17) for targeted (untargeted) attacks, which verifies that using $\ell_2$ in video attacks will not cause significant changes to the video.

\noindent\textbf{TI and AOA.} Drawing insights from TI results in Table~\ref{tab:attack_performance_ucf101} and Table~\ref{tab:attack_performance_hmdb51} in the supplement, it becomes evident that the PatchAttack demonstrates a significant deficiency in temporal consistency, which can be attributed to the incorporation of solid color patches. In contrast, several alternative methods showcase superior temporal consistency, with LogoStyleFool performing the best across the majority of cases. Attributed to its transparency, the AOA associated with the Adv-watermark is minimal. Furthermore, LogoStyleFool boasts the smallest AOA, resulting in the least amount of video occlusion when compared to the other two methods. The AOA of STDE~\cite{jiang2023efficient} (18.70\% in targeted attacks) is rather high that abundant semantic information is redacted.

\noindent\textbf{Analyses for Query Number.} Under the same black-box attack setting, the current video attacks~\cite{jiang2019black,cao2023stylefool} still require over $10^4$ of queries under global perturbations. In comparison, our attack considers regional perturbations based on logos whose semantics are also well-maintained. We argue that achieving an Average Query (AQ) of approximately 20,000 (targeted) and 2,000 (untargeted) is not only a commendable outcome but is also deemed acceptable, particularly in offline attack scenarios. In particular, we achieve an FR of over 80\% and an AQ of around 900 in untargeted attacks even when Stage 3 is not considered. It can be seen as a variant of our method, which can be used when the query budget is rather low.

\begin{figure*}[t]
\begin{center}
  \includegraphics[width=0.7\linewidth]{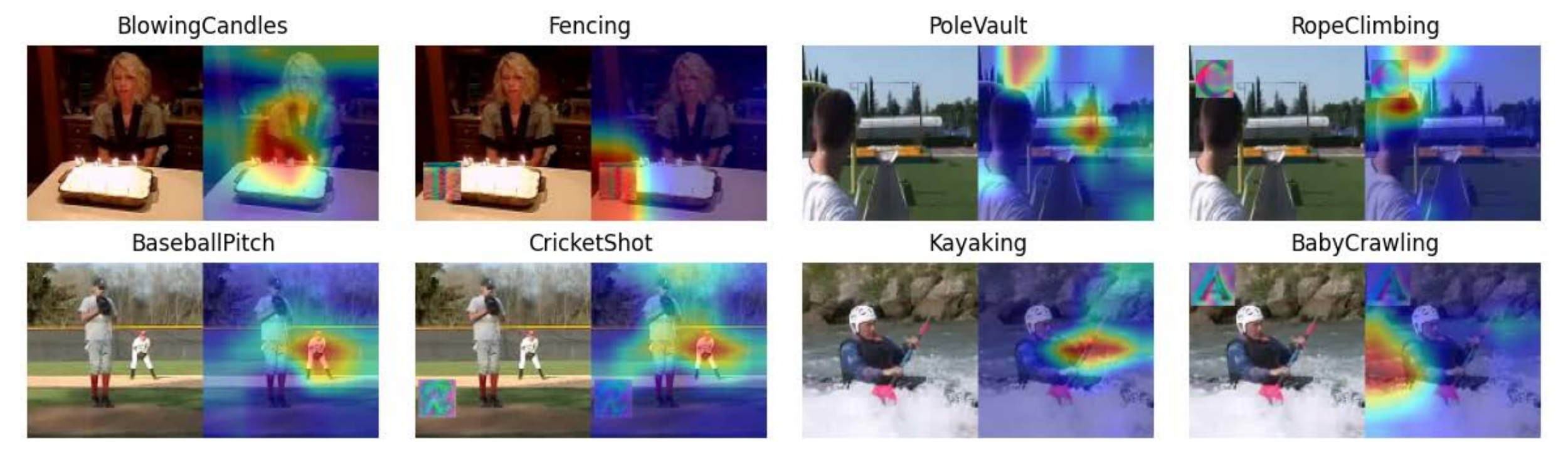}
\end{center}
  \caption{Grad-CAM visualizations of LogoStyleFool. Top row: targeted, bottom row: untargeted.}
\label{fig:grad_cam}
\end{figure*}

\noindent\textbf{Grad-CAM Visualizations.} To further evaluate the attack performance of LogoStyleFool, we use Grad-CAM~\cite{selvaraju2017grad} to visualize the local regions where the model focuses. The examples in Figure~\ref{fig:grad_cam} show that our optimized logos have a strong ability to mislead video recognition models. As a result, adding a stylized logo can move important regions from the semantic regions that are consistent with the original label to the logo area, or even other areas which do not overlap with the logo, since our attack generates adversarial examples once successfully misleading the video classifier and does not require high scores of the target class.

\subsection{Ablation Study}\label{subsec:ablation}
We verify the effectiveness of our proposed LogoStyleFool by considering different scenarios in each attacking stage. In Stage 1, we consider 1) randomly selecting style images (skipping Stage 1); 2) using solid color to initialize the style image instead of random initialization. In Stage 2, we consider 3) $\mu _a = 0$, \ie not penalizing the logo area; 4) $\mu _d = 0$, \ie not penalizing the distance from the logo to the corners. In Stage 3, we consider 5) only optimizing the perturbation one round (similar to SimBA~\cite{guo2019simple}); 6) optimizing one point per frame ($T$ points together) in a step. Table~\ref{tab:ablation} reports the results under different scenarios when attacking C3D on UCF-101. In scenarios 1) and 2), the style images are not prone to carry the information of the target class, leading to attack inefficiency in targeted attacks. We observe that due to the high likelihood that random style images are already adversarial, the average query drops a lot in untargeted attacks, which can be regarded as an improvement for LogoStyleFool. We also notice that although $^2$AQ is slightly lower than LogoStyleFool when the style image is initialized with solid color, Stage 3 consumes many queries. One possible reason is that the monotonous color in the optimized style image increases the difficulty of Stage 3. In scenarios 3) and 4), the FR improves a little, and the AQ reduces significantly in untargeted attacks. However, the larger logo area or its position near the center of the image results in the core action of the video being obstructed, thus affecting human visual perception. We denote average logo area as $\bar a = \sum\nolimits_i {k_i^2hw} $ and average minimum distance to the corners as $\bar d_m = \sum\nolimits_i {{d_{m,i}}} $ for all adversarial videos. Compared with $\bar a$ of 646.7(733.3) and $\bar d_m$ of 13.6(11.7) for targeted (untargeted) attacks in LogoStyleFool, $\bar a$ becomes 976.8(894.0) for targeted (untargeted) attacks in scenario 3), and $\bar d_m$ becomes 37.1(35.2) for targeted (untargeted) attacks in scenario 4). Considering the naturalness reduction, we do not adopt these two scenarios. We also consider different optimization strategies in scenarios 5) and 6). Untargeted attacks still exhibit better query performance than LogoStyleFool, as the perturbation required is minimal. For targeted attacks, a single round of optimization is not enough to move the video across the decision boundary, resulting in low FR. Since highly refined modification of each pixel value in the video is needed to achieve challenging targeted attacks, we attribute the loss of attack performance to the dimensionality increase from images to videos. Optimizing one pixel value per frame in each step leads to a higher query in Stage 3 since the simultaneous adjustment of multiple pixels ignores the mutual influence between these pixels. Taking into account both fooling rates and query efficiency, scenarios 1) and 5) can serve as variants of untargeted attacks, while the original LogostyleFool setting achieves higher attack efficiency for targeted attacks.

\begin{table}[t]
\centering
\footnotesize
\resizebox{0.9\linewidth}{!}{
\begin{tabular}{cccccccccc}
\toprule
\multirow{2}{*}[-0.5ex]{Model} & 
\multirow{2}{*}[-0.5ex]{Attack} & \multicolumn{2}{c}{UCF-101} & \multicolumn{2}{c}{HMDB-51} \\
\cmidrule(r){3-4}\cmidrule(r){5-6}
& & LGS & PC & LGS & PC \\
\midrule
\multirow{5}{*}{C3D} 
& Adv-watermark & \textbf{41\%} & 38\% & 44\% & 39\% \\
& PatchAttack & 39\% & 29\% & 42\% & 36\%  \\
& BSC & \textbf{41\%} & \textbf{42\%} & 42\% & 44\% \\
& LogoStyleFool-${\ell_\infty}$ & 40\% & \textbf{42\%} & \textbf{45\%} & 47\% \\
& LogoStyleFool-${\ell_2}$  & 40\% & 39\% & \textbf{45\%} & \textbf{48\%}  \\
\midrule
\multirow{5}{*}{I3D} 
& Adv-watermark & 38\% & 42\% & 39\% & 41\% \\
& PatchAttack & 41\% & 36\% & 35\% & 29\% \\
& BSC & 40\% & 43\% & 34\% & 36\% \\
& LogoStyleFool-${\ell_\infty}$  & 46\% & \textbf{55\%} & 44\% & \textbf{46\%}  \\
& LogoStyleFool-${\ell_2}$  & \textbf{47\%} & 52\% & \textbf{46\%} & 41\% \\
\bottomrule
\end{tabular}
}
\caption{Fooling rate ($\uparrow$) of defense performance.}
\label{tab:defense_performance}
\vspace{-3mm}
\end{table}

\subsection{Defense Performance}
We also evaluate the defense performance of our proposed LogoStyleFool against two state-of-the-art patch-based defense methods, Local Gradients Smoothing (LGS)~\cite{LGS_naseer2019local} and PatchCleanser (PC)~\cite{xiang2022patchcleanser}. LGS regularizes gradients in the estimated noisy region and inhibits the values of high activation regions caused by adversarial noise. PC uses two rounds of the pixel masking algorithm on the input image to remove the effects of adversarial patches and recover correct predictions. We extend LGS and PC to videos. We compare the performance of our approach with Adv-watermark, PatchAttack and BSC in terms of fooling rate against both defense methods, as shown in Table~\ref{tab:defense_performance}. The fooling rate is averaged on both targeted and untargeted attacks. Although watermarks are transparent and concealed, these two patch-based defenses still have a certain defensive effect on Adv-watermark. Owing to the poor performance of targeted attacks (<14\%), the FRs of Adv-watermark, PatchAttack and BSC reduce a lot. The performance of LogoStyleFool is on par with those of other competitors when encountering defenses. We argue that the performance is good enough since we additionally consider perturbation optimization which adds irregular perturbations to the logo, while the perturbations of PatchAttack and BSC are smoother since only RL is considered. If we add perturbation optimization after RL for PatchAttack and BSC, the perturbations added to the bullet-screen comments or solid RGB colors are obvious enough to trigger the alarm of detection methods with humans in the loop. We provide more analyses when Stage 3 is excluded for LogoStyleFool in the supplement. We notice that there is still a gap between current defenses and a plausible defensive mechanism.

\section{Conclusion}
In this paper, we study the vulnerability of video recognition systems and propose a novel attack framework LogoStyleFool. To address the naturalness reduction of style transfer-based in all pixels, we consider regional perturbations by adding a stylized logo to the corner of the video. We design a SimBA-based strategy to select style references and use reinforcement learning to search for the optimal logo, style image and location parameters. Next, a perturbation optimizer named LogoS-DCT iteratively optimizes the adversarial video, which mitigates the problem of limited search space of RL-based methods in targeted attacks. In addition, we prove the upper bounds on the perturbations, in both $\ell_\infty$ and $\ell_2$ norms. Comprehensive experiments show that LogoStyleFool can significantly improve the attack performance while preserving semantic information. Furthermore, LogoStyleFool parades its powerfulness and robustness against two existing defenses against patch attacks. Future work will focus on exploring potential defenses towards such subregional perturbations based on style transfer.

\section{Acknowledgments}
The authors thank anonymous reviewers for their feedback that helped improve the paper. This work was supported in part by the National Natural Science Foundation of China (61972219), the Overseas Research Cooperation Fund of Tsinghua Shenzhen International Graduate School (HW2021013). This work was also supported in part by facilities of Ping An Technology (Shenzhen) Co., Ltd., China, and CSIRO's Data61, Australia.

\bibliography{aaai24}

\begin{thebibliography}{46}
\providecommand{\natexlab}[1]{#1}

\bibitem[{Brown et~al.(2017)Brown, Man{\'e}, Roy, Abadi, and Gilmer}]{brown2017adversarial}
Brown, T.~B.; Man{\'e}, D.; Roy, A.; Abadi, M.; and Gilmer, J. 2017.
\newblock Adversarial patch.
\newblock \emph{Advances in Neural Information Processing Systems}.

\bibitem[{Cao et~al.(2023)Cao, Xiao, Sun, Wang, Xue, and Wen}]{cao2023stylefool}
Cao, Y.; Xiao, X.; Sun, R.; Wang, D.; Xue, M.; and Wen, S. 2023.
\newblock Stylefool: Fooling video classification systems via style transfer.
\newblock In \emph{2023 IEEE Symposium on Security and Privacy (SP)}, 1631--1648. IEEE.

\bibitem[{Carreira and Zisserman(2017)}]{carreira2017i3d}
Carreira, J.; and Zisserman, A. 2017.
\newblock Quo vadis, action recognition? a new model and the kinetics dataset.
\newblock In \emph{Proceedings of the IEEE/CVF Conference on Computer Vision and Pattern Recognition}, 6299--6308.

\bibitem[{Chen et~al.(2022)Chen, Wei, Chen, Wu, and Jiang}]{chen2022attacking}
Chen, K.; Wei, Z.; Chen, J.; Wu, Z.; and Jiang, Y.-G. 2022.
\newblock Attacking video recognition models with bullet-screen comments.
\newblock In \emph{Proceedings of the AAAI Conference on Artificial Intelligence}, volume~36, 312--320.

\bibitem[{Chen et~al.(2017)Chen, Liu, Shen, Yue, Cao, and Ma}]{chen2017deepcoder}
Chen, T.; Liu, H.; Shen, Q.; Yue, T.; Cao, X.; and Ma, Z. 2017.
\newblock Deepcoder: A deep neural network based video compression.
\newblock In \emph{2017 IEEE Visual Communications and Image Processing (VCIP)}, 1--4. IEEE.

\bibitem[{Chen et~al.(2021)Chen, Xie, Pang, He, and Tian}]{chen2021appending}
Chen, Z.; Xie, L.; Pang, S.; He, Y.; and Tian, Q. 2021.
\newblock Appending adversarial frames for universal video attack.
\newblock In \emph{Proceedings of the IEEE/CVF Winter Conference on Applications of Computer Vision}, 3199--3208.

\bibitem[{Chindaudom et~al.(2020)Chindaudom, Siritanawan, Sumongkayothin, and Kotani}]{chindaudom2020adversarialqr}
Chindaudom, A.; Siritanawan, P.; Sumongkayothin, K.; and Kotani, K. 2020.
\newblock AdversarialQR: An adversarial patch in QR code format.
\newblock In \emph{2020 Joint 9th International Conference on Informatics, Electronics \& Vision (ICIEV) and 2020 4th International Conference on Imaging, Vision \& Pattern Recognition (icIVPR)}, 1--6. IEEE.

\bibitem[{Croce et~al.(2022)Croce, Andriushchenko, Singh, Flammarion, and Hein}]{croce2022sparse}
Croce, F.; Andriushchenko, M.; Singh, N.~D.; Flammarion, N.; and Hein, M. 2022.
\newblock Sparse-rs: a versatile framework for query-efficient sparse black-box adversarial attacks.
\newblock In \emph{Proceedings of the AAAI Conference on Artificial Intelligence}, volume~36, 6437--6445.

\bibitem[{Gao et~al.(2023)Gao, Zheng, Yu, Shan, Ding, and Han}]{gao2023deep}
Gao, M.; Zheng, F.; Yu, J.~J.; Shan, C.; Ding, G.; and Han, J. 2023.
\newblock Deep learning for video object segmentation: a review.
\newblock \emph{Artificial Intelligence Review}, 56(1): 457--531.

\bibitem[{Gatys, Ecker, and Bethge(2015)}]{gatys2015a}
Gatys, L.~A.; Ecker, A.~S.; and Bethge, M. 2015.
\newblock A neural algorithm of artistic style.
\newblock \emph{arXiv preprint arXiv:1508.06576}.

\bibitem[{Gong et~al.(2023)Gong, Dong, Ma, Camtepe, Nepal, and Xu}]{gong2023stealthy}
Gong, H.; Dong, M.; Ma, S.; Camtepe, S.; Nepal, S.; and Xu, C. 2023.
\newblock Stealthy Physical Masked Face Recognition Attack via Adversarial Style Optimization.
\newblock \emph{IEEE Transactions on Multimedia}.

\bibitem[{Goodfellow, Shlens, and Szegedy(2015)}]{goodfellow2015explaining}
Goodfellow, I.~J.; Shlens, J.; and Szegedy, C. 2015.
\newblock Explaining and harnessing adversarial examples.
\newblock In \emph{Proceedings of the International Conference on Learning Representations}.

\bibitem[{Guo et~al.(2019)Guo, Gardner, You, Wilson, and Weinberger}]{guo2019simple}
Guo, C.; Gardner, J.; You, Y.; Wilson, A.~G.; and Weinberger, K. 2019.
\newblock Simple black-box adversarial attacks.
\newblock In \emph{International Conference on Machine Learning}, 2484--2493. PMLR.

\bibitem[{Hu et~al.(2023)Hu, Jin, Zheng, Weng, and Ding}]{hu2023overview}
Hu, K.; Jin, J.; Zheng, F.; Weng, L.; and Ding, Y. 2023.
\newblock Overview of behavior recognition based on deep learning.
\newblock \emph{Artificial Intelligence Review}, 56(3): 1833--1865.

\bibitem[{Inkawhich et~al.(2018)Inkawhich, Inkawhich, Chen, and Li}]{inkawhich2018adversarial}
Inkawhich, N.; Inkawhich, M.; Chen, Y.; and Li, H. 2018.
\newblock Adversarial attacks for optical flow-based action recognition classifiers.
\newblock \emph{arXiv preprint arXiv:1811.11875}.

\bibitem[{Ji et~al.(2012)Ji, Xu, Yang, and Yu}]{ji20123d}
Ji, S.; Xu, W.; Yang, M.; and Yu, K. 2012.
\newblock 3D convolutional neural networks for human action recognition.
\newblock \emph{IEEE transactions on pattern analysis and machine intelligence}, 35(1): 221--231.

\bibitem[{Jia et~al.(2019)Jia, Wei, Cao, and Foroosh}]{jia2019comdefend}
Jia, X.; Wei, X.; Cao, X.; and Foroosh, H. 2019.
\newblock Comdefend: An efficient image compression model to defend adversarial examples.
\newblock In \emph{Proceedings of the IEEE/CVF Conference on Computer Vision and Pattern Recognition}, 6084--6092.

\bibitem[{Jia et~al.(2020)Jia, Wei, Cao, and Han}]{jia2020adv}
Jia, X.; Wei, X.; Cao, X.; and Han, X. 2020.
\newblock Adv-watermark: A novel watermark perturbation for adversarial examples.
\newblock In \emph{Proceedings of the 28th ACM International Conference on Multimedia}, 1579--1587.

\bibitem[{Jiang et~al.(2023)Jiang, Chen, Huang, Wang, Yang, Li, Wang, and Zhang}]{jiang2023efficient}
Jiang, K.; Chen, Z.; Huang, H.; Wang, J.; Yang, D.; Li, B.; Wang, Y.; and Zhang, W. 2023.
\newblock Efficient decision-based black-box patch attacks on video recognition.
\newblock In \emph{Proceedings of the IEEE/CVF International Conference on Computer Vision}, 4379--4389.

\bibitem[{Jiang et~al.(2019)Jiang, Ma, Chen, Bailey, and Jiang}]{jiang2019black}
Jiang, L.; Ma, X.; Chen, S.; Bailey, J.; and Jiang, Y.-G. 2019.
\newblock Black-box adversarial attacks on video recognition models.
\newblock In \emph{Proceedings of the 27th ACM International Conference on Multimedia}, 864--872.

\bibitem[{Justin, Alexandre, and Li(2016)}]{justin2016perceptual}
Justin, J.; Alexandre, A.; and Li, F. 2016.
\newblock Perceptual Losses for Real-Time Style Transfer and Super-Resolution.
\newblock In \emph{Proceedings of the European Conference on Computer Vision (ECCV)}.

\bibitem[{Karmon, Zoran, and Goldberg(2018)}]{karmon2018lavan}
Karmon, D.; Zoran, D.; and Goldberg, Y. 2018.
\newblock Lavan: Localized and visible adversarial noise.
\newblock In \emph{International Conference on Machine Learning}, 2507--2515. PMLR.

\bibitem[{Kuehne et~al.(2011)Kuehne, Jhuang, Garrote, Poggio, and Serre}]{kuehne2011hmdb}
Kuehne, H.; Jhuang, H.; Garrote, E.; Poggio, T.; and Serre, T. 2011.
\newblock HMDB: a large video database for human motion recognition.
\newblock In \emph{Proceedings of the IEEE/CVF International Conference on Computer Vision (ICCV)}, 2556--2563. IEEE.

\bibitem[{Lei, Xing, and Chen(2020)}]{lei2020blind}
Lei, C.; Xing, Y.; and Chen, Q. 2020.
\newblock Blind video temporal consistency via deep video prior.
\newblock \emph{Advances in Neural Information Processing Systems}, 33: 1083--1093.

\bibitem[{Li, Schmidt, and Kolter(2019)}]{camera_li2019adversarial}
Li, J.; Schmidt, F.; and Kolter, Z. 2019.
\newblock Adversarial camera stickers: A physical camera-based attack on deep learning systems.
\newblock In \emph{International Conference on Machine Learning}, 3896--3904. PMLR.

\bibitem[{Li et~al.(2019)Li, Neupane, Paul, Song, Krishnamurthy, Roy-Chowdhury, and Swami}]{li2019stealthy}
Li, S.; Neupane, A.; Paul, S.; Song, C.; Krishnamurthy, S.~V.; Roy-Chowdhury, A.~K.; and Swami, A. 2019.
\newblock Stealthy Adversarial Perturbations Against Real-Time Video Classification Systems.
\newblock In \emph{Proceedings of the Symposium on Network and Distributed Systems Security (NDSS)}.

\bibitem[{Li et~al.(2018)Li, Liu, Kautz, and Yang}]{li2018learning}
Li, X.; Liu, S.; Kautz, J.; and Yang, M.-H. 2018.
\newblock Learning linear transformations for fast arbitrary style transfer.
\newblock \emph{arXiv preprint arXiv:1808.04537}.

\bibitem[{Ma et~al.(2019)Ma, Zhang, Jia, Zhao, Wang, and Wang}]{ma2019image}
Ma, S.; Zhang, X.; Jia, C.; Zhao, Z.; Wang, S.; and Wang, S. 2019.
\newblock Image and video compression with neural networks: A review.
\newblock \emph{IEEE Transactions on Circuits and Systems for Video Technology}, 30(6): 1683--1698.

\bibitem[{Naseer, Khan, and Porikli(2019)}]{LGS_naseer2019local}
Naseer, M.; Khan, S.; and Porikli, F. 2019.
\newblock Local gradients smoothing: Defense against localized adversarial attacks.
\newblock In \emph{2019 IEEE Winter Conference on Applications of Computer Vision (WACV)}, 1300--1307. IEEE.

\bibitem[{Pony, Naeh, and Mannor(2021)}]{pony2021flickering}
Pony, R.; Naeh, I.; and Mannor, S. 2021.
\newblock Over-the-Air Adversarial Flickering Attacks Against Video Recognition Networks.
\newblock In \emph{Proceedings of the IEEE/CVF Conference on Computer Vision and Pattern Recognition}.

\bibitem[{Sage et~al.(2018)Sage, Agustsson, Timofte, and Van~Gool}]{sage2018logo}
Sage, A.; Agustsson, E.; Timofte, R.; and Van~Gool, L. 2018.
\newblock Logo synthesis and manipulation with clustered generative adversarial networks.
\newblock In \emph{Proceedings of the IEEE Conference on Computer Vision and Pattern Recognition}, 5879--5888.

\bibitem[{Selvaraju et~al.(2017)Selvaraju, Cogswell, Das, Vedantam, Parikh, and Batra}]{selvaraju2017grad}
Selvaraju, R.~R.; Cogswell, M.; Das, A.; Vedantam, R.; Parikh, D.; and Batra, D. 2017.
\newblock Grad-cam: Visual explanations from deep networks via gradient-based localization.
\newblock In \emph{Proceedings of the IEEE international conference on computer vision}, 618--626.

\bibitem[{Soomro, Zamir, and Shah(2012)}]{soomro2012ucf101}
Soomro, K.; Zamir, A.~R.; and Shah, M. 2012.
\newblock UCF101: A dataset of 101 human actions classes from videos in the wild.
\newblock \emph{arXiv preprint arXiv:1212.0402}.

\bibitem[{Szegedy et~al.(2014)Szegedy, Zaremba, Sutskever, Bruna, Erhan, Goodfellow, and Fergus}]{szegedy2014intriguing}
Szegedy, C.; Zaremba, W.; Sutskever, I.; Bruna, J.; Erhan, D.; Goodfellow, I.; and Fergus, R. 2014.
\newblock Intriguing properties of neural networks.
\newblock In \emph{Proceedings of the International Conference on Learning Representations}.

\bibitem[{Tran et~al.(2015)Tran, Bourdev, Fergus, Torresani, and Paluri}]{tran2015learning}
Tran, D.; Bourdev, L.; Fergus, R.; Torresani, L.; and Paluri, M. 2015.
\newblock Learning spatiotemporal features with 3d convolutional networks.
\newblock In \emph{Proceedings of the IEEE/CVF International Conference on Computer Vision (ICCV)}, 4489--4497.

\bibitem[{Wang, Sha, and Yang(2021)}]{wang2021reinforcement}
Wang, Z.; Sha, C.; and Yang, S. 2021.
\newblock Reinforcement learning based sparse black-box adversarial attack on video recognition models.
\newblock In \emph{Proceedings of International Joint Conference on Artificial Intelligence}.

\bibitem[{Wei, Wang, and Yan(2023)}]{wei2023efficient}
Wei, X.; Wang, S.; and Yan, H. 2023.
\newblock Efficient Robustness Assessment Via Adversarial Spatial-Temporal Focus on Videos.
\newblock \emph{IEEE Transactions on Pattern Analysis and Machine Intelligence}.

\bibitem[{Wei, Yan, and Li(2022)}]{wei2022sparse}
Wei, X.; Yan, H.; and Li, B. 2022.
\newblock Sparse black-box video attack with reinforcement learning.
\newblock \emph{International Journal of Computer Vision}, 130(6): 1459--1473.

\bibitem[{Wei et~al.(2019)Wei, Zhu, Yuan, and Su}]{wei2019sparse}
Wei, X.; Zhu, J.; Yuan, S.; and Su, H. 2019.
\newblock Sparse adversarial perturbations for videos.
\newblock In \emph{Proceedings of the AAAI Conference on Artificial Intelligence (AAAI)}, volume~33, 8973--8980.

\bibitem[{Wei et~al.(2020)Wei, Chen, Wei, Jiang, Chua, Zhou, and Jiang}]{wei2020heuristic}
Wei, Z.; Chen, J.; Wei, X.; Jiang, L.; Chua, T.-S.; Zhou, F.; and Jiang, Y.-G. 2020.
\newblock Heuristic black-box adversarial attacks on video recognition models.
\newblock In \emph{Proceedings of the AAAI Conference on Artificial Intelligence (AAAI)}, volume~34, 12338--12345.

\bibitem[{Williams(1992)}]{williams1992simple}
Williams, R.~J. 1992.
\newblock Simple statistical gradient-following algorithms for connectionist reinforcement learning.
\newblock \emph{Reinforcement learning}, 5--32.

\bibitem[{Xiang, Mahloujifar, and Mittal(2022)}]{xiang2022patchcleanser}
Xiang, C.; Mahloujifar, S.; and Mittal, P. 2022.
\newblock $\{$PatchCleanser$\}$: Certifiably Robust Defense against Adversarial Patches for Any Image Classifier.
\newblock In \emph{31st USENIX Security Symposium (USENIX Security 22)}, 2065--2082.

\bibitem[{Xiao et~al.(2019)Xiao, Deng, Li, Lee, Edwards, Yi, Song, Liu, and Molloy}]{xiao2019advit}
Xiao, C.; Deng, R.; Li, B.; Lee, T.; Edwards, B.; Yi, J.; Song, D.; Liu, M.; and Molloy, I. 2019.
\newblock Advit: Adversarial frames identifier based on temporal consistency in videos.
\newblock In \emph{Proceedings of the IEEE/CVF International Conference on Computer Vision (ICCV)}, 3968--3977.

\bibitem[{Yan and Wei(2021)}]{yan2021efficient}
Yan, H.; and Wei, X. 2021.
\newblock Efficient sparse attacks on videos using reinforcement learning.
\newblock In \emph{Proceedings of the 29th ACM International Conference on Multimedia}, 2326--2334.

\bibitem[{Yang et~al.(2020)Yang, Kortylewski, Xie, Cao, and Yuille}]{yang2020patchattack}
Yang, C.; Kortylewski, A.; Xie, C.; Cao, Y.; and Yuille, A. 2020.
\newblock Patchattack: A black-box texture-based attack with reinforcement learning.
\newblock In \emph{Proceedings of the European Conference on Computer Vision (ECCV)}.

\bibitem[{Zhou et~al.(2022)Zhou, Porikli, Crandall, Van~Gool, and Wang}]{zhou2022survey}
Zhou, T.; Porikli, F.; Crandall, D.; Van~Gool, L.; and Wang, W. 2022.
\newblock A survey on deep learning technique for video segmentation.
\newblock \emph{IEEE Transactions on pattern analysis and machine intelligence}, 1--20.

\end{thebibliography}

\onecolumn
\newpage

\appendix

\section{LogoStyleFool: Vitiating Video Recognition Systems via Logo Style Transfer\\(Supplementary Material)}

\subsection{Pseudo Code for LogoS-DCT}
Algorithm~\ref{alg:logos_dct} shows the concrete process of LogoS-DCT ($\logosdct$ in Algorithm~\ref{alg:logostylefool}), which optimizes the perturbation in the logo area until the whole video becomes adversarial. $\update$ represents the update process in a certain pixel. As mentioned previously, each pixel in each update process can only be changed by one of $\left\{ - \eta, 0, \eta \right\}$ compared to the initial pixel. The update process follows the direction of increasing target class confidence score in targeted attacks or decreasing the original class confidence score in untargeted attacks.

\begin{algorithm}[ht]
\caption{LogoS-DCT.}\label{alg:logos_dct}
\KwIn{Black-box classifier $f$, stylized video $x_s^*$, original label $y_0$, target label ${y_t}$, mask matrix $M$, orthogonal frequency set $Q_{DCT}$, step size $\eta$, perturbation threshold ${\varepsilon}$.}
\KwOut{Adversarial video ${x_{adv}}$.}
$x_{adv} \gets x_s^*$\;
$\vartheta  \gets p\left( {{y_t}|{x_{adv}}} \right)$ // replace $y_t$ with $y_0$ for untargeted \;
\While {$f \left( x_{adv}\right) \ne y_t$ // \rm{replace it with $f \left( x_{adv}\right) = y_0$ for untargeted} } {
    $\rm{p} \gets$ randomly sampled from $Q_{DCT}$\;
    \For {$\kappa \in \left\{ -1, 0, 1 \right\}$}{
        ${x_{temp}} \leftarrow \update\left( {x_s^*,M,\kappa ,\eta ,{\rm{p}}, \varepsilon } \right)$\;
        \If {$p\left( {{y_t}|{x_{temp}}} \right) > \vartheta$ // \rm{replace it with $p\left( {{y_0}|{x_{temp}}} \right) < \vartheta$ for untargeted} }{
            $x_{adv} \gets x_{temp}$\;
            $\vartheta  \gets p\left( {{y_t}|{x_{adv}}} \right)$ // replace $y_t$ with $y_0$ for untargeted \;
        }
    }
}
\rm{\textbf{return}} $x_{adv}$. 
\end{algorithm}

\subsection{Additional Experimental Details}
\noindent\textbf{Dataset Introduction.}
Collected from YouTube, UCF-101~\cite{soomro2012ucf101} is an action recognition dataset that contains 13,320 videos from 101 different classes, including sports actions, human-human interaction, \etc. All videos last for over 27 hours. HMDB-51~\cite{kuehne2011hmdb} is another video classification dataset collected from YouTube and Google. It contains 6,849 videos from 51 classes.

\noindent\textbf{Model Introduction.}
C3D uses 3D convolution to learn temporal features and achieves excellent classification accuracy. Focusing on the relationship between two consecutive frames, I3D distinguishes actions from the perspective of optical flow, and it also secures comparable classification performance.

\noindent\textbf{Metrics Introduction.}
We introduction of metrics we used in our attacks are as follows. 

1) Fooling rate (FR): the ratio of adversarial videos that can successfully fool the classifier to the target class (targeted attack) or any other class (untargeted attack) within the query limit. 

2) Average query (AQ): the average query number to succeed in finding adversarial videos. 
For untargeted attacks, we find only a few videos need to go through Stage 3, the query of which will increase the AQ to some extent. To exhibit more clearly the excellent performance of LogoStyleFool, especially in untargeted attacks, we additionally calculate $^2$FR and $^2$AQ to represent the fooling rate and the average query if only the first two stages are considered. We also note down the average query in each stage in the hyperparameter experiments and ablation study to better observe the attack efficiency. We denote them as AQ$_1$, AQ$_2$, and AQ$_3$.

3) Average Occluded Area (AOA): the average area percentage occluded by logos/patches in the video.

4) Temporal Inconsistency (TI): we use the warping error $E_{warp}$ in paper \cite{lei2020blind} to measure the temporal inconsistency. The warping error considers both short-term and long-term inconsistency and is calculated as follows.
\begin{equation}\label{epair}
{E_{pair}}\left( {{x_t},{x_s}} \right) = \frac{1}{{HWC}}{O_{t,s}}{\left\| {{x_t} - {\mathcal W}\left( {{x_s}} \right)} \right\|_1},
\end{equation}
\begin{equation}\label{ewarp}
{E_{warp}}\left( {{x_t}} \right) = \frac{1}{{T - 1}}\sum\nolimits_{t = 2}^T {{E_{pair}}\left( {{x_t},{x_1}} \right) + {E_{pair}}\left( {{x_t},{x_{t - 1}}} \right)},
\end{equation}
where $O_{t,s}$ represents the occlusion mask matrix for the $t$-th and $s$-th frames $x_t$ and $x_s$, and $\mathcal W$ signifies the backward warping procedure utilizing optical flow.

\noindent\textbf{Hyperparameter Settings.}
In order to obtain the best performance, we carry out a batch of grid searches on three important hyperparameters in LogoStyleFool, style image number $N_s$, logo number $N_l$, and step size $\eta$. We randomly select 20 videos from UCF-101 and conduct both targeted and untargeted attacks against C3D with different hyperparameters. We report the attacking results in Table~\ref{tab:parameter_styleimg_num}, Table~\ref{tab:parameter_logo_num} and Table~\ref{tab:parameter_step_size}. The style image number $N_s$ directly affects AQ$_1$: the larger the value of $N_s$, the more queries are executed in Stage 1. A smaller $N_s$ will result in insufficient search space and bring difficulty for attacks, while a larger $N_s$ will make for the redundancy of search space and also affect the attack efficiency. As for logo number $N_l$, AQ$_2$ tends to converge as $N_l$ increases. When $N_l = $ 50 or 100, AQ$_3$ in both targeted and untargeted attacks drops by a large margin compared to others. Additionally, we discover that after style transfer, logos with letter content typically look more natural. One tenable reason could be that the naturalness of letters is less likely to be lost compared to a complex pattern since letters are relatively simple in terms of outline and are usually single-colored. It is the step size $\eta$ that influences the attack efficiency most, since Stage 3 is the most time-consuming process. Theoretically, since we pull the samples back onto the $\ell_\infty$ ball after each update, larger or smaller step sizes can result in optimization falling into local optima. This speculation is consistent with the experimental results -- a moderate step size attains a higher attack efficiency. Therefore, we set style image number $N_s = 5$, logo number $N_l = 100$, and step size $\eta = 0.2$ for both targeted and untargeted attacks.

\begin{table*}[htbp]
\centering
\resizebox{0.8\linewidth}{!}{
\begin{tabular}{ccrrrrrrrrrrrrrrrr}
\toprule
\multirow{2}{*}[-0.5ex]{$N_s$} & \multicolumn{5}{c}{UCF-101-Targeted} & \multicolumn{5}{c}{UCF-101-Untargeted} \\
\cmidrule(r){2-6}\cmidrule(r){7-11}
& FR$\uparrow$ & AQ$_1$ & AQ$_2$ & AQ$_3$ & AQ$\downarrow$ & FR$\uparrow$ & AQ$_1$ & AQ$_2$ & AQ$_3$ & AQ$\downarrow$ \\
\midrule
1 & 40\% & 917.3 & 1,691.4 & 20,153.2 & 22,761.9 & 100\% & 1.0 & 512.7 & 1,664.7 & 2,178.4 \\
3 & 35\% & 1,742.0 & 1,655.7 & 18,347.1 & 21,744.8 & 100\% & 1.1 & 517.3 & 1,588.3 & 2,116.7 \\
5 & 55\% & 2,487.3 & 1,724.8 & 15,530.4 & 19,742.5 & 100\% & 1.1 & 520.4 & 1,066.2 & 1,587.7 \\
7 & 45\% & 6,975.4 & 1,721.3 & 18,872.7 & 27,569.4 & 100\% & 1.4 & 531.8 & 1,315.3 & 1,848.5 \\
9 & 40\% & 10,246.8 & 1,744.0 & 21,152.4 & 33,143.2 & 100\% & 1.2 & 511.9 & 1,281.0 & 1,794.1 \\
\bottomrule
\end{tabular}
}
\caption{Results of LogoStyleFool with various style image number $N_s$. Metric details are provided in the experimental setup.}
\label{tab:parameter_styleimg_num}
\end{table*}

\begin{table*}[h]  
\centering
\resizebox{0.82\linewidth}{!}{
\begin{tabular}{ccrrrrrrrrrrrrrrrr}
\toprule
\multirow{2}{*}[-0.5ex]{$N_l$} & \multicolumn{5}{c}{UCF-101-Targeted} & \multicolumn{5}{c}{UCF-101-Untargeted} \\
\cmidrule(r){2-6}\cmidrule(r){7-11}
& FR$\uparrow$ & AQ$_1$ & AQ$_2$ & AQ$_3$ & AQ$\downarrow$ & FR$\uparrow$ & AQ$_1$ & AQ$_2$ & AQ$_3$ & AQ$\downarrow$ \\
\midrule
10 & 40\% & \multirow{6}{*}[-0ex]{2,525.9$\pm$39.0} & 1,352.2 & 31,303.1 & 35,207.0 & 100\% & \multirow{6}{*}[-0ex]{1.2$\pm$0.2} & 495.8 & 1,871.7 & 2,369.4 \\
20 & 50\% &  & 1,649.0 & 27,918.7 & 32,049.0 & 100\% &  & 488.0 & 1,354.2 & 1,843.1 \\
50 & 55\% &  & 1,731.5 & 24,355.6 & 28,669.6 & 100\% &  & 502.5 & 911.8 & 1,414.5 \\
80 & 45\% &  & 1,744.8 & 25,883.1 & 30,116.9 & 100\% &  & 491.6 & 1,434.9 & 1,927.2 \\
100 & 55\% &  & 1,722.7 & 22,913.6 & 27,147.6 & 100\% &  & 417.5 & 771.5 & 1,189.2 \\
120 & 40\% &  & 1,752.1 & 26,171.8 & 30,463.4 & 100\% &  & 400.3 & 1,833.8 & 2,234.4 \\
\bottomrule
\end{tabular}
}
\caption{Results of LogoStyleFool with various logo number $N_l$. Metric details are provided in the experimental setup.}
\label{tab:parameter_logo_num}
\end{table*}

\begin{table*}[ht]  
\centering
\resizebox{0.82\linewidth}{!}{
\begin{tabular}{ccrrrrrrrrrrrrrrrr}
\toprule
\multirow{2}{*}[-0.5ex]{$\eta$} & \multicolumn{5}{c}{UCF-101-Targeted} & \multicolumn{5}{c}{UCF-101-Untargeted} \\
\cmidrule(r){2-6}\cmidrule(r){7-11}
& FR$\uparrow$ & AQ$_1$ & AQ$_2$ & AQ$_3$ & AQ$\downarrow$ & FR$\uparrow$ & AQ$_1$ & AQ$_2$ & AQ$_3$ & AQ$\downarrow$ \\
\midrule
0.05 & 20\% & \multirow{5}{*}[-0ex]{2,641.6$\pm$53.0} & \multirow{5}{*}[-0ex]{1,546.9$\pm$65.2} & 32,395.8 & 36,562.6 & 100\% & \multirow{5}{*}[-0ex]{1.2$\pm$0.2} & \multirow{5}{*}[-0ex]{461.7$\pm$19.0} & 6,861.5 & 7,342.9 \\
0.1 & 35\% &  &  & 26,006.3 & 30,099.6 & 100\% &  &  & 3,795.0 & 4,251.9 \\
0.2 & 50\% &  &  & 23,741.4 & 27,993.9 & 100\% &  &  & 792.8 & 1,226.2 \\
0.3 & 40\% &  &  & 27,816.2 & 32,027.2 & 100\% &  &  & 1,092.9 & 1,569.1 \\
0.4 & 40\% &  &  & 28,139.5 & 32,358.3 & 100\% &  &  & 1,331.5 & 1,796.7 \\
\bottomrule
\end{tabular}
}
\caption{Results of LogoStyleFool with various step size $\eta$. Metric details are provided in the experimental setup.}
\label{tab:parameter_step_size}
\end{table*}

\begin{table*}[t]
\centering
\footnotesize
\resizebox{0.85\linewidth}{!}{
\begin{tabular}{ccrrrrrrrrrrrrrrrr}
\toprule
\multirow{2}{*}[-0.5ex]{Model} & 
\multirow{2}{*}[-0.5ex]{Attack} & \multicolumn{4}{c}{HMDB-51-Targeted} & \multicolumn{4}{c}{HMDB-51-Untargeted}\\
\cmidrule(r){3-6}\cmidrule(r){7-10}
& & FR($^2$FR)$\uparrow$ & AQ($^2$AQ)$\downarrow$ & AOA$\downarrow$ & TI$\downarrow$ & FR($^2$FR)$\uparrow$ & AQ($^2$AQ)$\downarrow$ & AOA$\downarrow$ & TI$\downarrow$ \\
\midrule
\multirow{5}{*}{C3D} 
& Adv-watermark & 3\% & 732.7 & \textbf{4.37\%} & 5.81 & 53\% & 831.0 & \textbf{5.15\%} & 4.20 \\
& PatchAttack & 4\% & 36,622.3& 6.78\% & 143.54 & 78\% & 5,193.0 & 6.24\% & 274.53\\
& BSC & 15\% & 28,014.5 & 6.07\% & 3.87 & 88\% & 3,590.2 & 7.81\% & 4.33\\
& LogoStyleFool-${\ell_\infty}$ & 63\%(14\%) & \textbf{19,253.5}(2,117.4) & 5.24\% & \textbf{3.38} & 98\%(\textbf{90\%}) & \textbf{2,689.5}(\textbf{857.8}) & 5.72\% & 4.14\\
& LogoStyleFool-${\ell_2}$ & \textbf{73\%}(11\%) & 24,157.2(2,685.0) & 5.33\% & 3.39 & \textbf{100\%}(86\%) & 3,149.0(873.2) & 5.44\% & \textbf{4.13} \\
\midrule
\multirow{5}{*}{I3D} 
& Adv-watermark & 3\% & 632.5& \textbf{4.33\%} & 6.24 & 68\% & 137.8 & \textbf{4.72\%} & 4.11\\
& PatchAttack & 14\% & 29,218.4 & 5.98\% & 108.52 & 77\% & 1,118.5 & 5.54\% & 117.10\\
& BSC & 21\% & 30,204.7 & 8.15\% & 3.34 & 85\% & 1,305.4 & 6.92\% & 4.15\\
& LogoStyleFool-${\ell_\infty}$ & \textbf{69\%}(18\%) & \textbf{19,583.6}(3,035.7) & 5.57\% & \textbf{3.01} & \textbf{100\%}(\textbf{96\%}) & \textbf{925.9}(795.0) & 5.70\% & \textbf{3.48} \\
& LogoStyleFool-${\ell_2}$ & 64\%(16\%) & 24,834.1(4,528.3) & 5.62\% & 3.06 & \textbf{100\%}(94\%) & 1,081.8 (\textbf{727.5}) & 5.25\% & 3.60 \\
\bottomrule
\end{tabular}
}
\caption{Attack performance comparison on HMDB-51. Metric details are provided in the experimental setup.}
\label{tab:attack_performance_hmdb51}
\end{table*}

\noindent\textbf{Other Parameter Settings.}
For other parameters in LogoStyleFool, we set the batch size in reinforcement learning as 30 in untargeted attacks and 50 in targeted attacks, penalty coefficients $\mu_a = 0.004$ and $\mu_d = 0.2$, and the perturbation threshold $\varepsilon = 0.1$ under $\ell_\infty$ restrictions. Since the size of the logos in the LLD dataset is 32$\times $32, we set the logo height $h = 32$ and width $w=32$. The scaling ratio $k$ is chosen from $\left\{ 0.75,0.8125,0.875,0.9375,1.0 \right\}$. Values exceeding 1.0 are not considered as they may obscure the semantic part of the video. The videos are attacked using one RTX 3080 GPU. 

\begin{figure}[t]
    \centering
    \includegraphics[width=0.5\linewidth]{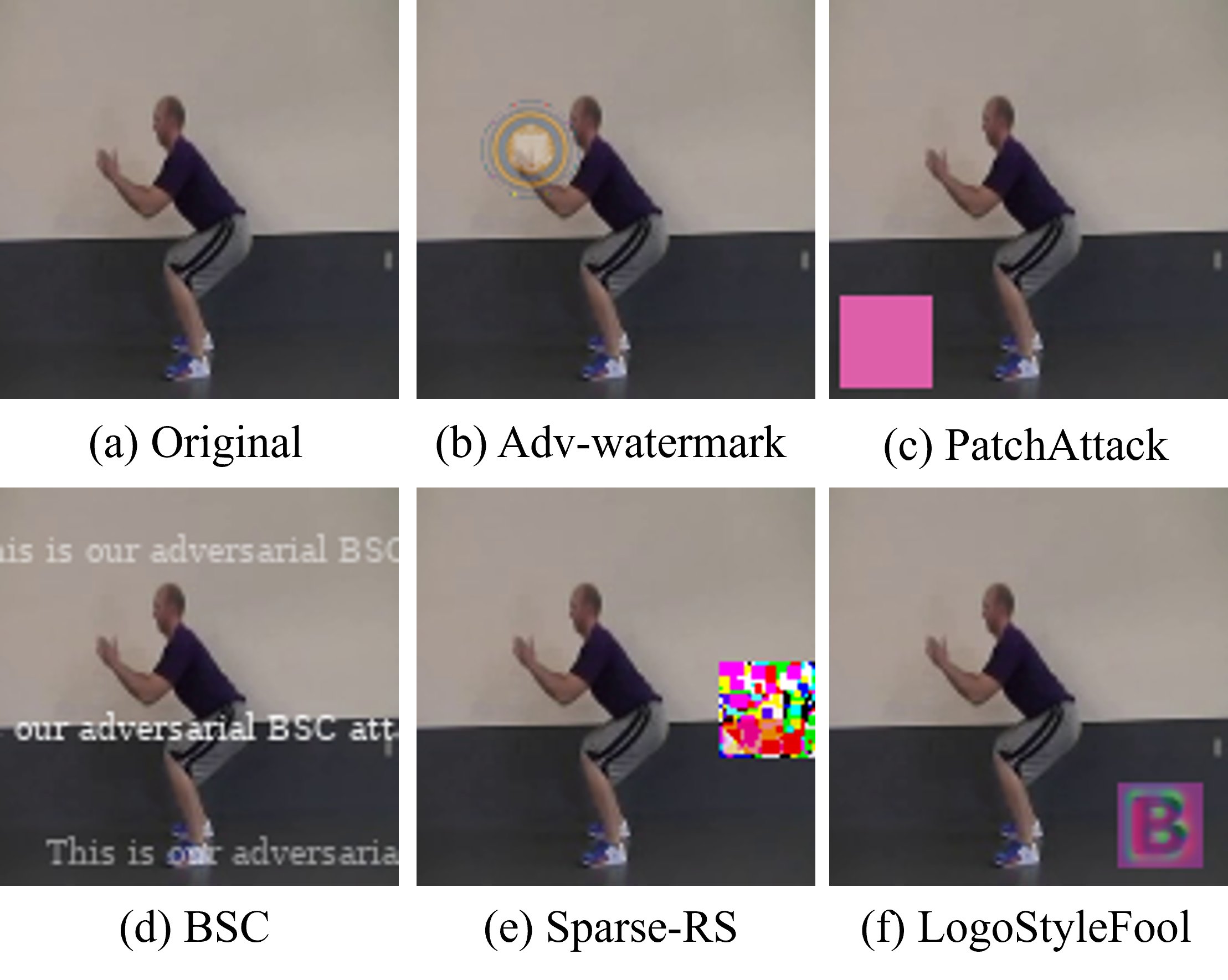}
    \caption{Examples of different attacks.}
    \label{fig:five_images}
\end{figure}

\noindent\textbf{Attack Visualizations.}
Figure~\ref{fig:five_images} shows the adversarial examples of five different attacks. We also involve an image attack, Sparse-RS~\cite{croce2022sparse}, which performs better in attack but exhibits the least naturalness. Visually, LogoStyleFool best preserves the semantic information of the videos.

\begin{table*}[htbp]  
\centering
\resizebox{0.9\linewidth}{!}{
\begin{tabular}{ccrrrrrrrrrrrrrrrr}
\toprule
\multirow{2}{*}[-0.5ex]{Attack scenario} & \multicolumn{5}{c}{UCF-101-Targeted} & \multicolumn{5}{c}{UCF-101-Untargeted} \\
\cmidrule(r){2-6}\cmidrule(r){7-11}
& FR($^2$FR)$\uparrow$ & AQ$_1$ & AQ$_2$ & AQ$_3$ & AQ($^2$AQ)$\downarrow$ & FR($^2$FR)$\uparrow$ & AQ$_1$ & AQ$_2$ & AQ$_3$ & AQ($^2$AQ)$\downarrow$ \\
\midrule
Random style image & 29\%(5\%) & 0 & 1,661.8 & 40,583.2 & 42,249.7(1,661.8) & 96\%(80\%) & 0 & 282.3 & 912.4 & 1,194.7(282.3) \\
Solid color initialization & 32\%(7\%) & 6,735.3 & 1,825.8 & 37,622.5 & 46,183.6(8,561.1) & 82\%(74\%) & 1.0 & 695.3 & 6,211.7 & 6,908.0(696.3) \\
\midrule
$\mu _a = 0$ & 55\%(17\%) & 2,574.0 & 1,255.8 & 20,724.2 & 24,554.0(3,829.8) & 98\%(91\%) & 1.1 & 286.3 & 250.7 & 538.1(287.4) \\
$\mu _d = 0$ & 53\%(19\%) & 2,611.5 & 1,361.5 & 25,112.8 & 29,085.8(3,973.0) & 97\%(93\%) & 1.2 & 192.4 & 612.6 & 505.2(192.6) \\
\midrule
One round & 27\%(12\%) & 2,607.2 & 1,539.1 & 9,810.8 & 13,957.1(4,146.3) & 97\%(76\%) & 1.1 & 395.2 & 462.5 & 858.8(396.3) \\
One point per frame & 19\%(11\%) & 2,542.3 & 1,607.1 & 53,382.9 & 57,532.3(4,149.4) & 96\%(90\%) & 1.2 & 383.5 & 1,355.7 & 1,740.4(384.7) \\
\bottomrule
\end{tabular}
}
\caption{Ablation results of LogoStyleFool. Metric details are provided in the experimental setup.}
\label{tab:ablation}
\end{table*}

\noindent\textbf{Analyses about Sparse-RS.}
We also adapt Sparse-RS~\cite{croce2022sparse} from images to videos. All parameters are set by default. After experiments under the same AOA, the FR (76\%) and AQ (4,273.4) are superior to ours, but the semantic information of the video is severely impacted due to using random color patches. As depicted in Figure~\ref{fig:five_images}, among the five attacks, Sparse-RS applies a disorganized patch onto the video, resulting in the least natural appearance. In contrast, LogoStyleFool demonstrates a plausible capacity in maintaining the natural visual coherence of the video content.

\noindent\textbf{Analyses about Defense.}
From the methodological perspective, the comparisons in Table~\ref{tab:defense_performance} are not entirely fair. This is because our competitors do not have the optimization stage in their pipelines. Specifically, Adv-watermark, PatchAttack and BSC do not consider perturbation optimization, resulting in poor attack performance (especially in targeted attacks) but relatively smoother perturbations that bypass the defenses. In contrast, Stage 3 of LogoStyleFool boosts its attack performance but hinders its evasion capability against defenses. On the other hand, it is infeasible to append such a stage to their methods since, by the nature of their design, there is no unified optimization stage applicable to all the methods. 
Hence, it is more fair to remove Stage 3 from LogoStyleFool in the comparison. We find that we can obtain an average FR of over 58\% in the defense experiment if Stage 3 is removed. The average FR of over 58\% is averaged from FRs of both untargeted and targeted attacks. However, the FR of targeted attacks is behind that of untargeted attacks. In untargeted attacks, most attacks succeed without requiring Stage 3, which brings a huge advantage in the defense experiments (over 66\% FR when Stage 3 is not considered). Therefore, when compared more fairly, our method has a significant advantage in circumventing defenses.

\subsection{Proofs}
We first provide some preliminary definitions for proofs.
\begin{definition}\label{definition1}
Given a state space $I = \left\{ { - 1,0,1} \right\}$, we define a stochastic chain $\zeta (t), t \in {T_{opt}} = \left\{ {0,1,...,n,...} \right\}$, which represents the sign direction in the $t$-th optimization round. 
\end{definition}

\begin{lemma}\label{lemma2}
$\zeta (t)$ is a homogeneous Markov chain.
\end{lemma}
\textbf{Proof of Lemma~\ref{lemma2}.}
The pixel value is modified in the direction that increases the score of the target class. Since the direction change in pixel value only depends on the value of the pixel in the previous round and is independent of the directions of all previous rounds, $\zeta (t)$ therefore satisfies Markov property, \ie $\forall {t_0},{t_1},...,{t_n} \in {T_{opt}}$, ${t_0} < {t_1} < ..., < {t_n}$, $\forall {i_0},{i_1},...,{i_n} \in I$, we have
\begin{equation}
\Pr \left\{ {\zeta ({t_{n + 1}}) = {i_{n + 1}}|\zeta ({t_0}) = {i_0},...,\zeta ({t_n}) = {i_n}} \right\} = \Pr \left\{ {\zeta ({t_{n + 1}}) = {i_{n + 1}}|\zeta ({t_n}) = {i_n}} \right\}.
\end{equation}
Obviously, the state transition probability is independent of time $t$. Thus, $\zeta (t)$ is a homogeneous Markov chain.

Based on the Markov assumption, we can derive the rest of the theorems.

\begin{definition}\label{definition2}
Define another stochastic chain $\xi (t), t \in {T_{opt}} = \left\{ {0,1,...,n,...} \right\}$ which satisfies $\xi (t) = 0$ if $\zeta ({t_n}) = 0$, $\xi (t) = \prod\limits_{j = n - k + 1}^n {\zeta ({t_j})} $, $k = 1,2,...,n$ if $\zeta ({t_n}) \ne 0,\zeta ({t_{n - 1}}) \ne 0,...\zeta ({t_{n - k + 1}}) \ne 0$, and $\zeta ({t_{n - k}}) = 0$.
\end{definition}

\begin{lemma}\label{lemma3}
$\xi (t)$ is also a Markov chain.
\end{lemma}
\textbf{Proof of Lemma~\ref{lemma3}.}
Based on Lemma~\ref{lemma2} and Definition~\ref{definition2}, this is obvious since the value of $\xi (t+1)$ also only depends on $\xi (t)$.

\noindent\textbf{Proofs of Proposition~\ref{proposition1} and Theorem~\ref{theorem1}.}
For expressive simplicity, we use $\xi _{m,t}$ to denote the value $\xi (t)$ for the $m$-th pixel in the $t$-th optimization round. Based on Lemma~\ref{lemma3}, we can only use $\xi _{m,t}$ in the last optimization round to describe the perturbation ${\delta _K} $.
If $K > d$, we have
\begin{equation}
\begin{aligned}
{\delta _K} 
 &= \sum\limits_{m = 1}^{K - \left\lfloor {\frac{K}{d}} \right\rfloor d} {{\rm{clip}}_{ - \varepsilon }^{ + \varepsilon }\left( {{A^T}\left( {{\xi _{m,\left\lfloor {\frac{K}{d}} \right\rfloor  + 1}}\eta {{\rm{p}}_m}} \right)A} \right)}  + \sum\limits_{m = K - \left\lfloor {\frac{K}{d}} \right\rfloor d + 1}^d {{\rm{clip}}_{ - \varepsilon }^{ + \varepsilon }\left( {{A^T}\left( {{\xi _{m,\left\lfloor {\frac{K}{d}} \right\rfloor }}\eta {{\rm{p}}_m}} \right)A} \right)} \\
 &= \sum\limits_{m = 1}^d {{\rm{clip}}_{ - \varepsilon }^{ + \varepsilon }\left( {{A^T}\left( {{\gamma _m}\eta {{\rm{p}}_m}} \right)A} \right)} ,
\end{aligned}
\end{equation}
where ${\gamma _m} \in \left\{ { -1,0,1 } \right\}$ denotes the final direction for the $m$-th pixel, which can be expressed as

\begin{equation}
{\gamma _m} = \left\{ {\begin{array}{*{20}{l}}
{{\xi _{m,\left\lfloor {\frac{K}{d}} \right\rfloor  + 1}},1 \le m \le K - \left\lfloor {\frac{K}{d}} \right\rfloor d,}\\
{{\xi _{m,\left\lfloor {\frac{K}{d}} \right\rfloor }},K - \left\lfloor {\frac{K}{d}} \right\rfloor d < m \le d.}
\end{array}} \right.
\end{equation}

If $K < d$, we have
\begin{equation}
{\delta _K} = \sum\limits_{m = 1}^K {{\rm{clip}}_{ - \varepsilon }^{ + \varepsilon }\left( {{A^T}\left( {{\xi _{m,1}}{{\rm{p}}_m}} \right)A} \right)}  = \sum\limits_{m = 1}^K {{\rm{clip}}_{ - \varepsilon }^{ + \varepsilon }\left( {{A^T}\left( {{\gamma _m}\eta {{\rm{p}}_m}} \right)A} \right)} .
\end{equation}
Combining the above two equations, the perturbation after $K$ steps can be expressed as
\begin{equation}
{\delta _K} = \sum\limits_{m = 1}^{\min \left\{ {K,d} \right\}} {{\rm{clip}}_{ - \varepsilon }^{ + \varepsilon }\left( {{A^T}\left( {{\gamma _m}\eta {{\rm{p}}_m}} \right)A} \right)} .
\end{equation}
Therefore, Proposition~\ref{proposition1} is proved. Considering $M$ is only a mask matrix with 0 and 1, and the perturbation is clipped by $\varepsilon$ in each step, the final perturbation is bounded by
\begin{equation}
{\left\| {M \odot {\delta _K}} \right\|_\infty} \le \varepsilon \rho \sqrt {\min \left\{ {K,d} \right\}} .
\end{equation}
Henceforth, the proof of Theorem~\ref{theorem1} is concluded.

\noindent\textbf{Proof of Lemma~\ref{lemma1}.}
Obviously, $\left\| {{A_{0*}}} \right\|_2^2 = \sum\limits_{j = 0}^{d - 1} {{c^2}\left( 0 \right)}  = 1$. For $i \in \left\{ {1,2,...,d - 1} \right\}$, we have 
\begin{equation}
\left\| {{A_{i*}}} \right\|_2^2 = \sum\limits_{j = 0}^{d - 1} {{c^2}\left( i \right){{\cos }^2}\left[ {\frac{{\left( {j + 0.5} \right)\pi }}{d}i} \right]}  = \frac{d}{2}{c^2}\left( i \right) + \frac{1}{2}\sum\limits_{j = 0}^{d - 1} {\cos \left[ {\frac{{\left( {2j + 1} \right)\pi }}{d}i} \right]}  = 1.
\end{equation}
The last equality follows from the Euler's formula. Thus, for any row in matrix A, $\left\| {{A_{i*}}} \right\|_2 = 1$.

\noindent\textbf{Proof of Theorem~\ref{theorem2}.}
Since ${\rm{p}}_m$ is orthogonal, we have
\begin{equation}
\begin{aligned}
{\left\| {{\delta _K}} \right\|_2^2} &= \left\| {\sum\limits_{m = 1}^{\min \left\{ {K,d} \right\}} {{A^T}\left( {{\gamma _m}\eta {{\rm{p}}_m}} \right)A} } \right\|_2^2\\
 &= \sum\limits_{m = 1}^{\min \left\{ {K,d} \right\}} {\left\| {{A^T}\left( {{\gamma _m}\eta {{\rm{p}}_m}} \right)A} \right\|_2^2}  + \sum\limits_{m = 1}^{\min \left\{ {K,d} \right\}} {\sum\limits_{{m^{'}} = 1,{m^{'}} \ne m}^{\min \left\{ {K,d} \right\}} {{{\left\| {{{\left[ {{A^T}\left( {{\gamma _m}\eta {{\rm{p}}_m}} \right)A} \right]}^T}{A^T}\left( {{\gamma _{{m^{'}}}}\eta {{\rm{p}}_{{m^{'}}}}} \right)A} \right\|}_2}} } \\
 &= \sum\limits_{m = 1}^{\min \left\{ {K,d} \right\}} {\left\| {{A^T}\left( {{\gamma _m}\eta {{\rm{p}}_m}} \right)A} \right\|_2^2}  + \sum\limits_{m = 1}^{\min \left\{ {K,d} \right\}} {\sum\limits_{{m^{'}} = 1,{m^{'}} \ne m}^{\min \left\{ {K,d} \right\}} {{\eta ^2}{{\left\| {{\gamma _m}{\gamma _{{m^{'}}}}{A^T}{\rm{p}}_m^T{{\rm{p}}_{{m^{'}}}}A} \right\|}_2}} } \\
 &= \sum\limits_{m = 1}^{\min \left\{ {K,d} \right\}} {{{\left\| {{{\left[ {{A^T}\left( {{\gamma _m}\eta {{\rm{p}}_m}} \right)A} \right]}^T}\left[ {{A^T}\left( {{\gamma _m}\eta {{\rm{p}}_m}} \right)A} \right]} \right\|}_2}} \\
 &= \sum\limits_{m = 1}^{\min \left\{ {K,d} \right\}} {\gamma _m^2{\eta ^2}{{\left\| {{A^T}\left( {{\rm{p}}_m^T{{\rm{p}}_m}} \right)A} \right\|}_2}} \\
 &= \sum\limits_{m = 1}^{\min \left\{ {K,d} \right\}} {\gamma _m^2{\eta ^2}\left\| {{{\rm{p}}_m}A} \right\|_2^2} \\
 &= \sum\limits_{m = 1}^{\min \left\{ {K,d} \right\}} {\gamma _m^2{\eta ^2}\left\| {{A_{i*}}} \right\|_2^2} \\
 &= \sum\limits_{m = 1}^{\min \left\{ {K,d} \right\}} {\gamma _m^2{\eta ^2}} .
\end{aligned}
\end{equation}

Considering $M$ is only a mask matrix with 0 and 1 and $\gamma _m^2 \le 1$, we have
\begin{equation}
{\left\| {M \odot {\delta _K}} \right\|_2} \le \eta \rho \sqrt {\min \left\{ {K,d} \right\}}.
\end{equation}
Therefore, the proof ends here.

\end{document}